%% file: main.tex
\documentclass[lettersize,journal]{IEEEtran}
\usepackage{amsmath,amsfonts}
\usepackage{algorithmic}
\usepackage{array}
\usepackage{textcomp}
\usepackage{stfloats}
\usepackage{url}
\usepackage{verbatim}
\usepackage{graphicx}
\def\BibTeX{{\rm B\kern-.05em{\sc i\kern-.025em b}\kern-.08em
    T\kern-.1667em\lower.7ex\hbox{E}\kern-.125emX}}
\usepackage{balance}

\usepackage{times}
\usepackage{soul}
\usepackage{url}
\usepackage[hidelinks]{hyperref}
\usepackage[utf8]{inputenc}
\usepackage[small]{caption}
\usepackage{graphicx}
\usepackage{amsmath}
\usepackage{amsthm}
\usepackage{booktabs}
\usepackage{algorithm}
\usepackage{algorithmic}
\usepackage[switch]{lineno}

\usepackage{xspace}
\newcommand{\name}{\texttt{MoLS}\xspace}
\usepackage{subcaption}
\usepackage{multirow}
\usepackage{amssymb}
\input{mathcommand}

\usepackage{soul}  
\usepackage[table]{xcolor}
\usepackage{bm}
\definecolor{codeblue}{RGB}{0, 82, 147}     
\definecolor{codegreen}{RGB}{0, 128, 0}    
\definecolor{codegray}{RGB}{100, 100, 100}  
\definecolor{codeorange}{RGB}{230, 145, 56} 
\definecolor{darkerblue}{rgb}{0,0.08,0.45} 
\definecolor{royalblue}{RGB}{65,105,225}
\definecolor{lightblue}{RGB}{221,235,247}
\definecolor{fig3blue}{RGB}{47, 122, 232}  
\definecolor{fig3red}{RGB}{213, 32, 52}
\definecolor{fig3green}{RGB}{0, 137, 72} 
\definecolor{fig3yellow}{RGB}{217, 161, 5}
\definecolor{gray94}{gray}{.94}
\definecolor{gray90}{gray}{.90}

\newcommand{\red}[1]{\textcolor{red}{#1}}

\definecolor{darkgreen}{RGB}{34,139,34}
\definecolor{lightgreen}{RGB}{50, 160, 83}
\newcommand{\green}[1]{\textcolor{lightgreen}{#1}}
\newcommand{\gray}[1]{\textcolor{gray}{#1}}
\newcommand{\gbf}[1]{\green{\bf{#1}}}
\newcommand{\rbf}[1]{\red{\bf{#1}}}

\newcolumntype{g}{>{\columncolor{gray94}}c} 
\newcolumntype{b}{>{\columncolor{lightblue}}c} 
\newcommand{\grow}[1]{\rowcolor{gray94}{#1}} 
\newcommand{\brow}[1]{\rowcolor{lightblue}{#1}} 

\usepackage{amsmath,amssymb,amsthm}
\makeatletter
\@ifundefined{molsassumption}{\newtheorem{molsassumption}{Assumption}}{}
\@ifundefined{molslemma}{\newtheorem{molslemma}{Lemma}}{}
\@ifundefined{molstheorem}{\newtheorem{molstheorem}{Theorem}}{}
\@ifundefined{molscorollary}
{\newtheorem{molscorollary}{Corollary}}{}
\@ifundefined{molsproposition}{\newtheorem{molsproposition}{Proposition}}{}
\makeatother

\begin{document}

\title{MoLS: SNR-Guided Module-Wise Learning-Rate Calibration for Large Language Models}

\author{Ziqing Wen, Zhouyang Liu, Jiahuan Wang, Ping Luo, Junlin Zeng, Li Shen, Dongsheng Li, and Tao Sun$^*$

\thanks{Manuscript received.} \thanks{Corresponding author: Tao Sun.}
\thanks{Ziqing Wen, Zhouyang Liu, Jiahuan Wang, Ping Luo, Junlin Zeng, Tao Sun, Dongsheng Li are with the National Key Laboratory of Parallel and Distributed Computing, National University of Defense Technology, Changsha, 410073, China (e-mail: 
\href{zqwen@nudt.edu.cn}{zqwen@nudt.edu.cn}; 
\href{liuzhouyang20@nudt.edu.cn}{liuzhouyang20@nudt.edu.cn};
\href{wangjiahuan@nudt.edu.cn}{wangjiahuan@nudt.edu.cn};
\href{luoping@nudt.edu.cn}{luoping@nudt.edu.cn}; 
\href{zengjunlin21@nudt.edu.cn}{zengjunlin21@nudt.edu.cn}
\href{dsli@nudt.edu.cn}{dsli@nudt.edu.cn};
\href{suntao.saltfish@outlook.com}{suntao.saltfish@outlook.com}).}
\thanks{Li Shen is with the School of Computer and Key Laboratory of Advanced Microprocessors Chips and Systems, National University
of Defense Technology, Changsha, 410073, China (e-mail: \href{lishen@nudt.edu.cn}{lishen@nudt.edu.cn}).}
}

\markboth{Journal of \LaTeX\ Class Files,~Vol.~18, No.~9, September~2020}%
{How to Use the IEEEtran \LaTeX \ Templates}

\maketitle

\begin{abstract}
Large language models (LLMs) contain functionally distinct modules
with heterogeneous gradient statistics. Although Adam(W) adapts
updates at the parameter level, selecting relative learning rates
across modules often relies on costly manual tuning.
We study this problem through gradient signal-to-noise ratio (SNR).
Our analysis identifies structural sources of module-wise SNR
differences and relates SNR to the attenuation of Adam's normalized
mean-gradient component.
Based on these findings, we propose
\textbf{Module-wise Learning Rate Scaling via SNR (\name)},
a lightweight method that replaces manual multiplier search with
one-shot calibration.
\name{} estimates module-level SNRs after an initial adaptation
period and uses inverse-square-root scaling to determine relative
learning-rate multipliers.
The multipliers are introduced smoothly during warmup and then
held fixed while the global learning-rate schedule continues.
Our theoretical results quantify the effect of calibration and
its sensitivity to estimation error and relative SNR drift.
Extensive experiments across LLM pre-training and fine-tuning
show that \name{} reduces perplexity and improves downstream
accuracy with minimal computational overhead.
It achieves performance comparable to carefully tuned module-specific
learning rates and integrates with memory-efficient optimizers
and parameter-efficient fine-tuning.
\end{abstract}


\input{Sections/main_body}

\bibliography{reference}
\bibliographystyle{IEEEtran}
\input{Sections/appendix}

\end{document}

%% file: mathcommand.tex
\usepackage{amsmath,amsfonts,bm}

\def\eqref#1{equation~\ref{#1}}

\def\1{\bm{1}}

\def\rvg{{\mathbf{g}}}

\def\rvm{{\mathbf{m}}}

\def\rvv{{\mathbf{v}}}
\def\rvw{{\mathbf{w}}}

\DeclareMathAlphabet{\mathsfit}{\encodingdefault}{\sfdefault}{m}{sl}
\SetMathAlphabet{\mathsfit}{bold}{\encodingdefault}{\sfdefault}{bx}{n}



%% file: Sections/main_body.tex
\section{Introduction}

\IEEEPARstart{L}{arge} Language Models (LLMs) contain functionally
distinct modules, including self-attention, feed-forward networks,
embeddings, and output heads~\cite{vaswani2017attention,radford2019languagegpt2,touvron2023llama}.
These components serve different roles and exhibit different gradient
and optimization properties~\cite{zhang2024transformers_need_adam,liu2020understanding_difficult,ormaniec2024doestransformer}.
Adam and AdamW~\cite{kingma2014adam,loshchilov2017decoupled_adamw}
adapt parameter updates through element-wise moment estimates.
This adaptivity addresses individual gradient statistics, while the
choice of relative learning rates across functional modules remains
separate.

Recent studies report differences in gradient scale, curvature, and
optimization behavior across Transformer modules~\cite{kunstner2024heavy,zhang2025adammini}.
Zeng et al.~\cite{zeng2022glm} and Wang et al.~\cite{wang2025sharpness}
show the value of module-specific learning rates.
The practical challenge is choosing their relative values.
Searching over module-wise multipliers adds tuning cost, especially
when moving between model sizes and training tasks.

We study this problem through gradient signal-to-noise ratio (SNR).
Under a local moment approximation, Adam's normalized mean-gradient
magnitude depends on the balance between the gradient mean and
variance~\cite{orvieto2025in_search_adam}.
Low SNR attenuates this mean component.
Our structural analysis identifies how input participation,
class-specific output parameters, and module derivative directions
contribute to SNR differences.
Measurements on language models reveal pronounced differences among
functional groups, motivating calibration from their gradient
statistics.

We propose \textbf{Module-wise Learning Rate Scaling via SNR
(\name)}, a lightweight method that converts early gradient statistics
into module-specific learning-rate multipliers.
After a short initial phase with a shared learning rate, \name{}
estimates module SNRs once and gradually introduces the resulting
multipliers over the remaining scheduled warmup.
The multipliers reach their targets at the original warmup endpoint
and remain fixed during the subsequent decay.
This procedure complements Adam's element-wise adaptivity with
module-wise calibration. It avoids a separate multiplier search
and requires no additional warmup stage.
Our contributions are as follows:

\begin{itemize}
    \item \textbf{Analysis of module-wise gradient SNR.}
    We identify structural sources of SNR heterogeneity and relate
    SNR to Adam's normalized mean-gradient signal.
    We quantify the effect of calibration, its sensitivity to
    relative SNR drift, and the consistency of finite-sample estimates.

    \item \textbf{SNR-guided learning-rate calibration.}
    We introduce a one-shot procedure that assigns module-specific
    learning-rate multipliers from early gradient statistics.
    It replaces manual multiplier search with calibration and uses the
    existing warmup interval to introduce the multipliers gradually.
    The original warmup length and the base optimizer's moment
    updates are preserved.

    \item \textbf{Experimental evaluation.}
    We evaluate \name{} in pre-training, supervised fine-tuning,
    and parameter-efficient fine-tuning.
    It improves the underlying optimizers across pre-training model
    scales up to 7B parameters and improves downstream accuracy.
    Additional studies establish its low overhead, compatibility with
    Adam-mini~\cite{zhang2025adammini} and LoRA~\cite{hu2022lora},
    and low sensitivity of final performance to the calibration budget.
    A structured search comparison shows performance close to the
    best manually tuned module-wise configuration.
\end{itemize}

\section{Related Works}

\paragraph{Efficient LLM Optimizers}
Adaptive optimizers such as Adam(W)~\cite{kingma2014adam,loshchilov2017decoupled_adamw}
are widely used in LLM training.
Research on optimizer performance and memory efficiency includes
Nesterov-style momentum~\cite{dozat2016incorporating_nadamw},
LAMB~\cite{You2020Large_lamb}, and parameter-noise--normalized
SGD~\cite{xu2024no_sgdgal}.
Other approaches use matrix-based preconditioners~\cite{jordan2024muon,vyas2024soap},
powered gradients~\cite{wang2025gradpower}, or clustering-based
learning rates~\cite{li2025taming_sgg}.
Block-shared normalizers~\cite{zhang2025adammini} and low-rank
projection methods~\cite{zhao2024galore,zhu2024apollosgdlikememoryadamwlevel}
address optimizer memory use.
\name{} complements these approaches by calibrating learning rates
at the functional-module level.

\paragraph{Heterogeneity of LLM modules}
LLM modules exhibit distinct optimization properties.
Zeng et al.~\cite{zeng2022glm} identified differences in gradient scale
between embeddings and other modules.
Zhang et al.~\cite{zhang2024transformers_need_adam} analyzed
block-diagonal Hessian structure and variation in module curvature.
Ormaniec et al.~\cite{ormaniec2024doestransformer} examined Hessian
blocks for the Query/Key and Value/Output projections, revealing
differences within attention.
Wang et al.~\cite{wang2025sharpness} analyzed module-wise sharpness
and introduced blockwise learning rates.
These studies motivate separate learning rates but also raise the
cost of selecting module-specific values.

\paragraph{Positioning of \name}
\name{} complements adaptive optimization with module-wise
learning-rate calibration.
Its SNR analysis links module gradient statistics to the
mean-gradient signal retained by Adam's normalization.
This relation motivates setting relative learning rates from
measured statistics, replacing a separate module-wise search
with one-shot calibration while preserving Adam's moment updates.

\section{Motivation}
We examine how SNR affects Adam's normalized mean-gradient component
and how module structure contributes to SNR differences.
We then connect this analysis to the language-model measurements
that motivate \textbf{Module-wise Learning Rate Scaling via SNR
(\name)}.

\begin{figure*}[!t]
    \centering
    \begin{subfigure}[b]{1.0\linewidth}
        \centering
        \includegraphics[width=\linewidth]{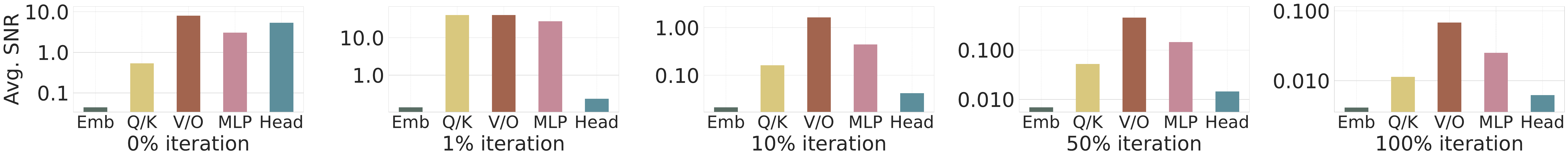}
        \caption{Pre-training LLaMA-60M on C4.}
    \end{subfigure}
    \begin{subfigure}[b]{1.0\linewidth}
        \centering
        \includegraphics[width=\linewidth]{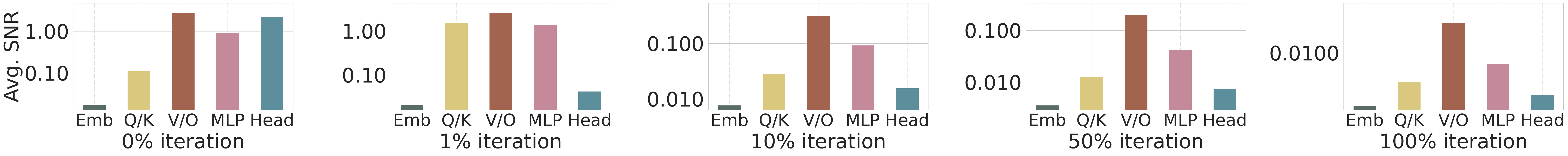}
        \caption{Pre-training LLaMA-130M on C4.}
    \end{subfigure}
    \caption{\textbf{SNR variation across modules during training.}
    The absolute SNR values change across training checkpoints,
    while differences among functional modules remain visible at
    both model scales. VO generally exhibits relatively high SNR,
    whereas embeddings and the output head often have lower SNR
    after the initial transient.}
    \label{fig:snr_c4_60m_130m}
\end{figure*}

\subsection{SNR-Dependent Signal Attenuation in Adam}

Adam and AdamW~\cite{kingma2014adam,loshchilov2017decoupled_adamw}
maintain element-wise first- and second-moment estimates of stochastic
gradients. Omitting moment bias correction and decoupled weight decay
in the following analysis, the moment recursions are
\begin{equation}\label{eq:adam_mv_update}
\begin{aligned}
    \rvm^{t}&=\beta_1\rvm^{t-1}+(1-\beta_1)\rvg^{t},\\
    \rvv^{t}&=\beta_2\rvv^{t-1}+(1-\beta_2)(\rvg^{t})^{\odot2},
\end{aligned}
\end{equation}
where $\beta_1,\beta_2\in[0,1)$ are decay rates and $\odot2$
denotes element-wise squaring. The corresponding update is
\begin{equation}\label{eq:adam_update}
    \rvw^{t+1}=\rvw^t-\eta_t\frac{\rvm^t}{\sqrt{\rvv^t}+\epsilon},
\end{equation}
where $\eta_t>0$ is the learning rate and $\epsilon>0$ is a
numerical stabilizer. All normalization operations are element-wise.

Let $\boldsymbol\mu=\mathbb E[\rvg]$ denote the local gradient mean
and $\boldsymbol\sigma^{\odot2}$ its coordinate-wise variance.
The moments in Eq.~(\ref{eq:adam_mv_update}) track the mean and raw
second moment, with
\begin{equation}
\nonumber
    \mathbb E[\rvg^{\odot2}]
    =\boldsymbol\mu^{\odot2}+\boldsymbol\sigma^{\odot2}.
\end{equation}
Adam's normalization therefore reflects both the mean gradient signal
and its stochastic variability.

For coordinate $i$ in group $m$, let $\mathcal D_{m,i}$ be the
magnitude of the expected normalized update.
We use a local stationary-moment approximation, replacing the moment
estimates by their population values and neglecting the stabilizer:
\begin{equation}\label{eq:adam_step_snr}
\begin{aligned}
    \mathcal D_{m,i}
    &:=\left|\mathbb E\left[
       \frac{m_{m,i}^t}{\sqrt{v_{m,i}^t}+\epsilon}\right]\right|
      \approx\frac{|\mu_{m,i}|}
                   {\sqrt{\mu_{m,i}^2+\sigma_{m,i}^2}}\\
    &=\sqrt{\frac{s_{m,i}}{1+s_{m,i}}},
\end{aligned}
\end{equation}
where $s_{m,i}=\mu_{m,i}^2/\sigma_{m,i}^2$ is the coordinate SNR
and $\sigma_{m,i}$ is the gradient standard deviation.
The approximation replaces the random numerator and denominator
with their population-moment counterparts.
The absolute value isolates the mean-gradient magnitude.

\paragraph{High-SNR regime}
As $s_{m,i}\to\infty$, the approximate magnitude approaches $1$,
and its signed counterpart approaches $\operatorname{sign}(\mu_{m,i})$,
connecting Adam's normalization to sign-based
updates~\cite{bernstein2018signsgd,kunstner2023noise_signsgd}.

\paragraph{Low-SNR regime}
When $\sigma_{m,i}\gg|\mu_{m,i}|$, Eq.~(\ref{eq:adam_step_snr})
gives $\mathcal D_{m,i}\approx\sqrt{s_{m,i}}$.
Low SNR therefore attenuates the normalized mean-gradient signal.
Coordinates with different SNRs can have different signal magnitudes
under a common learning rate.

\subsection{Structural Sources of Module-wise SNR Differences}
\label{sec:mols_structural}

Consider a decoder with an untied input embedding
$\mathbf E\in\mathbb R^{V\times d}$ and linear output head
$\mathbf U\in\mathbb R^{V\times d}$. For a context $\mathbf x$,
\begin{equation}
\nonumber
\begin{aligned}
 \mathbf h&=\mathcal T_{\boldsymbol\theta}(\mathbf E[\mathbf x]),
 &\mathbf z&=\mathbf U\mathbf h,\\
 \mathbf p&=\operatorname{softmax}(\mathbf z),
 &\ell(\mathbf x,y)&=-\log p_y.
\end{aligned}
\end{equation}
We use the same five groups as the calibration procedure: Emb, QK,
VO, MLP, and Head. QK contains query/key projection tensors, and
VO contains value/output projection tensors across layers.

\paragraph{Local setting and notation}
Fix a parameter state $\mathbf w^t$ and a context $\mathbf x$.
Assume the decoder is differentiable and the data label probabilities
$q_j=\mathbb P(y=j\mid\mathbf x)$ are positive.
These conditions permit differentiation and make label covariance
positive on zero-sum contrasts. Set
$\mathbf r=\mathbf p-\mathbf q$,
$\mathbf C=\operatorname{diag}(\mathbf q)-\mathbf q\mathbf q^\top$, and
$\Gamma=\mathbf r^\top\mathbf C^\dagger\mathbf r>0$.
For each coordinate, define the logit derivative
$\mathbf b_{m,i}=\partial\mathbf z/\partial w_{m,i}$.
Let $\kappa_m$ be the tensor-wise weighted mean of squared cosines
between $\mathbf C^{1/2}\mathbf b_{m,i}$ and
$(\mathbf C^\dagger)^{1/2}\mathbf r$.
Assign zero to zero-gradient coordinates.
The coefficient $0\leq\kappa_m\leq1$ measures the retention of mean
residual signal relative to label variability.
Structural scores use raw population moments without an SNR stabilizer.

\begin{molsproposition}[Structural decomposition of module SNR]
\label{prop:mols_structure}
Let $K=K(\mathbf x)$ be the number of distinct input indices and
$S_m(\mathbf x)$ the tensor-wise aggregated module score computed
from gradient moments conditional on $\mathbf x$. If the hidden coordinates are nonzero, then, for
$m\in\{\mathrm{QK},\mathrm{VO},\mathrm{MLP}\}$,
\begin{equation}
\begin{aligned}
 S_m(\mathbf x)&=\Gamma\kappa_m,\\
 S_{\mathrm{Emb}}(\mathbf x)&=\Gamma\frac K V\kappa_{E,\mathrm{act}},\\
 S_{\mathrm{Head}}(\mathbf x)&=\frac1V\sum_{j=1}^V
                \frac{r_j^2}{q_j(1-q_j)},
\end{aligned}
\label{eq:mols_structural_scores}
\end{equation}
where $\kappa_{E,\mathrm{act}}$ averages over the appearing embedding
rows. For $S_{\mathrm{Emb}}(\mathbf x)>0$,
\begin{equation}
\nonumber
 \frac{S_m(\mathbf x)}{S_{\mathrm{Emb}}(\mathbf x)}
 =\frac V K\frac{\kappa_m}{\kappa_{E,\mathrm{act}}}.
\end{equation}
With $q_{\max}=\max_jq_j$, the head comparison satisfies
\begin{equation}
\nonumber
 V(1-q_{\max})\kappa_m
 \leq\frac{S_m(\mathbf x)}{S_{\mathrm{Head}}(\mathbf x)}
 \leq V\kappa_m.
\end{equation}
\end{molsproposition}

Proposition~\ref{prop:mols_structure} separates three sources of
module SNR differences: input participation, class-specific output
parameters, and module derivative directions.
The internal-to-embedding ratio contains the structural factor $V/K$
and the directional factor $\kappa_m/\kappa_{E,\mathrm{act}}$.
QK and VO have distinct coefficients because they affect attention
weights and transformed values through different derivative paths.
The proof is in Appendix~\ref{app:mols_structural_proof}.

Lemmas~\ref{lem:mols_occurrence} and~\ref{lem:mols_context_mixing}
extend the analysis to token occurrence and aggregation across
contexts.
Module SNRs depend jointly on architecture, current parameters,
and data statistics.
We therefore estimate the scores from language-model training
gradients and use the measured ratios for calibration.

\subsection{Observed Heterogeneity and SNR-Guided Calibration}

We average coordinate-wise SNR estimates within each parameter
tensor and then across tensors in each functional group.
Figure~\ref{fig:snr_c4_60m_130m} shows pronounced differences among
the five groups on C4.
Figure~\ref{fig:snr_gpt_open} extends these measurements to other
model and dataset configurations.
Table~\ref{tab:alpha_value} shows distinct calibration factors across
model scales.

\paragraph{Early dynamics and calibration timing}
The measurements show a marked initial change in the head's relative
SNR. Clear differences among functional groups are already visible
at the 1\% checkpoint and persist later in training, while the
absolute SNR values continue to change.
We therefore collect calibration statistics after approximately
the first 1\% of the planned training steps, within the original
warmup interval. The remaining warmup steps gradually introduce
the module-specific learning rates.
Related work also observes high initial SNR followed by a decline
during GPT training~\cite{kosson2024warmup}.
Declining SNR reflects a weaker mean gradient relative to its
stochastic variability.

\paragraph{Module-wise rescaling}
Equation~(\ref{eq:adam_step_snr}) describes the mean-gradient signal
retained by Adam's normalization.
In the low-SNR regime, its magnitude scales with $\sqrt{s_{m,i}}$.
A common learning rate therefore leaves low-SNR coordinates with
a more attenuated mean component.
This motivates inverse-square-root module scaling to compensate
for that attenuation while retaining Adam's coordinate-wise adaptation:
\begin{equation}\label{eq:alpha_factor}
 \alpha_m=\sqrt{\frac{S_{\mathrm{base}}}{S_m}}.
\end{equation}
Here $S_{\mathrm{base}}=S_{m_{\mathrm{base}}}$ is the reference score,
so $\alpha_{\mathrm{base}}=1$. In the implementation, we substitute
the empirical scores $\widehat S_m$ and denote the resulting fixed
multipliers by $\widehat\alpha_m$. VO serves as the reference in
the reported LLaMA experiments.

This rule uses measured module ratios to compensate for
SNR-dependent mean-signal attenuation while preserving Adam's
moment recursions.
The target multipliers are estimated once. We introduce them
gradually over the remaining scheduled warmup to avoid an abrupt
learning-rate change at calibration, then keep them fixed during
decay.
Section~\ref{sec:mols_theory} analyzes the resulting signal disparity,
relative SNR drift, and finite-sample estimation.
Table~\ref{tab:snr_samples_ppl} evaluates the calibration budget,
and Algorithm~\ref{alg:snr_lr_simple} summarizes the procedure.

\paragraph{Comparison with existing work}
Wang et al.~\cite{wang2025sharpness} analyze sharpness heterogeneity
and introduce blockwise learning rates.
Li et al.~\cite{li2025taming_sgg} use dynamic gradient grouping and
group-specific scaling.
Kosson et al.~\cite{kosson2024warmup} relate SNR to early
representation changes and study SNR-dependent update scaling for
warmup.
\name{} instead uses a one-shot estimate after initial adaptation
to set fixed multipliers for functional groups.
It replaces a separate module-wise search with calibration while
retaining the base optimizer.

\section{Methodology}

\textbf{Module-wise Learning Rate Scaling via SNR (\name)}
calibrates relative learning rates for a chosen partition of trainable
parameters using early gradient statistics.
The method targets module-wise differences in normalized mean-gradient
attenuation while keeping calibration lightweight and compatible
with the existing training schedule.

\paragraph{Design considerations}
The design balances calibration cost, early adaptation, and a gradual
change in update scale.
Repeated SNR collection adds computation and data movement.
We therefore use one-shot estimation to capture functional differences
with a single statistics-collection stage.
Figures~\ref{fig:snr_c4_60m_130m} and~\ref{fig:snr_gpt_open}
show an initial SNR transient followed by persistent differences
among module types, motivating estimation after a short initial
training period.
The global schedule controls the overall step size, while the
calibrated multipliers specify its relative allocation across modules.
We introduce these multipliers through the remaining warmup to
avoid an abrupt change in module learning rates.

\paragraph{Functional grouping}
For LLaMA, we group parameters into Emb, QK, VO, MLP, and Head.
This partition follows their computational roles and the structural
distinctions in Proposition~\ref{prop:mols_structure}.
QK and VO receive separate factors because they affect attention
weights and transformed values through different derivative paths.
Table~\ref{tab:ablation_module} supports this separation.
Tensors with the same functional role share a multiplier across
layers, keeping the number of calibrated factors at five regardless
of model depth.

\paragraph{Calibration timing and one-shot SNR estimation}
Let $T$ denote the planned number of optimizer updates and
$T_{\mathrm w}$ the original warmup length.
The default calibration point is $t_0\approx0.01T$, with
$t_0<T_{\mathrm w}$.
All groups follow the same base learning-rate schedule until $t_0$.
This places calibration after a short shared-rate adaptation period
and leaves the remaining warmup for introducing module-specific rates.
At $\mathbf w^{t_0}$, we hold the parameters fixed and collect
$N$ stochastic mini-batch gradients.
Holding the state fixed makes these observations describe gradient
variability at a common parameter state.
We estimate coordinate SNRs from sample means and variances,
then average within each tensor and across tensors in a group
to obtain $\widehat S_m$.
The default pre-training budget is $N=32$ mini-batches of 512 examples,
or 16,384 examples in total.
Collection leaves the optimizer-update count and learning-rate
schedule unchanged.
Appendix~\ref{app:snr_details} gives the estimator details.

\paragraph{SNR-guided relative scaling}
For reference group $b=m_{\mathrm{base}}$, we apply the
inverse-square-root correction in Eq.~(\ref{eq:alpha_factor})
to the estimated module scores:
\begin{equation}
\nonumber
 \widehat\alpha_m=\sqrt{\frac{\widehat S_b}{\widehat S_m}}.
\end{equation}
Groups with lower SNR scores receive larger multipliers to compensate
for stronger mean-signal attenuation.
We use VO as the reference in the LLaMA experiments and set
$\widehat\alpha_b=1$.
Reference normalization fixes the common scale, while the measured
score ratios determine the relative learning-rate allocation.
The target multipliers are computed once and retained for the
remainder of training.

\paragraph{Integration with the existing warmup schedule}
Applying the target multipliers immediately would introduce a jump
in module learning rates.
We therefore transition each applied multiplier from $1$ to
$\widehat\alpha_m$ over the original interval from $t_0$ to
$T_{\mathrm w}$.
All multipliers reach their targets at the original warmup endpoint.
For $t\geq T_{\mathrm w}$, module $m$ follows
\begin{equation}
\nonumber
 \eta_{m,t}=\widehat\alpha_m\eta_t,
\end{equation}
where $\eta_t$ follows the original decay schedule.
This preserves the warmup length and subsequent decay, so no
additional warmup stage is needed.
The multipliers act through parameter-group learning rates in
Eq.~(\ref{eq:adam_update}), leaving Adam's moment recursions unchanged.
Section~\ref{sec:exp_efficiency} reports end-to-end overhead,
and Table~\ref{tab:snr_samples_ppl} evaluates the calibration budget.

\section{Theoretical Analysis}
\label{sec:mols_theory}

We analyze the scaling rule and how estimation error and later SNR
changes affect one-shot calibration.
Theorem~\ref{thm:mols_calibration} bounds signal disparity under the
population-moment model in Eq.~(\ref{eq:adam_step_snr}).
Theorem~\ref{thm:mols_one_shot} gives an exact relation between
relative score errors and multiplier errors, then separates
estimation error from subsequent SNR drift.
Corollary~\ref{cor:mols_finite_sample} establishes finite-sample
consistency under local sampling conditions.
Proofs are given in Appendix~\ref{app:mols_proofs}.

\begin{molsassumption}[Local moments]
\label{ass:mols_moments}
At each considered parameter state, gradient coordinates have finite
means and finite positive variances. Each analyzed module has a
positive population SNR score.
\end{molsassumption}
These conditions ensure well-defined SNR ratios at each parameter
state. The moments may vary throughout training.

\paragraph{Connection to module-wise learning rates}
Let $S_m^t$ be the population module score and $\widehat S_m$ its
estimate at calibration state $t_0$. For
$s_{m,i}^t=(\mu_{m,i}^t)^2/(\sigma_{m,i}^t)^2$, use fixed weights
$\omega_{m,i}>0$, $\sum_i\omega_{m,i}=1$, matching the method's
tensor-wise aggregation, and define
\begin{equation}
\nonumber
\begin{aligned}
 S_m^t&=\sum_i\omega_{m,i}s_{m,i}^t,\qquad
 (R_m^t)^2=\sum_i\omega_{m,i}\frac{s_{m,i}^t}{1+s_{m,i}^t},\\
 \tau_m^t&=\frac{\sum_i\omega_{m,i}(s_{m,i}^t)^2}{S_m^t}.
\end{aligned}
\end{equation}
In the moment model, $R_m^t$ is the weighted root-mean-square
normalized mean-gradient signal. The quantity $\tau_m^t$ combines
SNR scale and within-module heterogeneity.
At the same parameter state, multiplying the learning rate by $a_m$
changes the signal scale from $\eta_tR_m^t$ to $\eta_ta_mR_m^t$.
For nonempty $\mathcal G\subseteq\mathcal M$ and positive
$\mathbf a$, define
$\mathcal K_t(\mathbf a)=
\max_{m\in\mathcal G}a_mR_m^t/
\min_{m\in\mathcal G}a_mR_m^t$.
The common $\eta_t$ cancels, and a smaller $\mathcal K_t$ indicates
less module-wise signal disparity.

\paragraph{Effect of inverse-square-root scaling}
We first apply the scaling rule to population scores. The resulting
time-indexed factors are references for analysis.

\begin{molstheorem}[Residual signal disparity]
\label{thm:mols_calibration}
Under Assumption~\ref{ass:mols_moments}, let
$\alpha_m^t=\sqrt{S_b^t/S_m^t}$, $b=m_{\mathrm{base}}$, and
$\mathbf a^t=(\alpha_m^t)_m$. Then
\begin{equation}
 \sqrt{\frac{S_b^t}{1+\tau_m^t}}
 \leq\alpha_m^tR_m^t
 \leq\sqrt{\frac{S_b^t}{1+S_m^t}}.
\label{eq:mols_calibrated_sandwich}
\end{equation}
Set $S_-^t=\min_{m\in\mathcal G}S_m^t$,
$S_+^t=\max_{m\in\mathcal G}S_m^t$,
$\tau_+^t=\max_{m\in\mathcal G}\tau_m^t$, and
$H^t=\sqrt{(1+\tau_+^t)/(1+S_-^t)}$. Then
\begin{equation}
\nonumber
 \mathcal K_t(\mathbf a^t)\leq H^t,\qquad
 \frac{\mathcal K_t(\mathbf a^t)}{\mathcal K_t(\mathbf 1)}
 \leq\frac{(H^t)^2}{\sqrt{S_+^t/S_-^t}}.
\end{equation}
Thus, calibration strictly reduces signal disparity if
$S_+^t/S_-^t>(H^t)^4$.
\end{molstheorem}

Theorem~\ref{thm:mols_calibration} quantifies the compensation
provided by inverse-square-root scaling.
It replaces the leading module-specific SNR factor in the signal
bounds with a common reference factor.
Nonlinear saturation and within-module heterogeneity determine the
remaining disparity. The stated sufficient condition guarantees
that calibration reduces this disparity.

\paragraph{Accuracy of one-shot calibration}
The implementation uses fixed
$\widehat\alpha_m=\sqrt{\widehat S_b/\widehat S_m}$, so after warmup
$\eta_{m,t}/\eta_{n,t}=\widehat\alpha_m/\widehat\alpha_n$.
For positive vectors, define
\begin{equation}
\nonumber
 d_{\mathrm{rel}}(\mathbf x,\mathbf y)
 =\max_{m,n}\left|\log\frac{x_m/x_n}{y_m/y_n}\right|.
\end{equation}
Write $\mathbf s^t=(S_m^t)_m$,
$\widehat{\mathbf s}=(\widehat S_m)_m$, and
$\widehat{\mathbf a}=(\widehat\alpha_m)_m$.

\begin{molstheorem}[One-shot calibration]
\label{thm:mols_one_shot}
For positive scores and estimates,
\begin{equation}
\begin{aligned}
 \Delta^t&:=d_{\mathrm{rel}}(\widehat{\mathbf a},\mathbf a^t)
 =\tfrac12d_{\mathrm{rel}}(\widehat{\mathbf s},\mathbf s^t)\\
 &\leq\tfrac12d_{\mathrm{rel}}(\widehat{\mathbf s},\mathbf s^{t_0})
 +\tfrac12d_{\mathrm{rel}}(\mathbf s^{t_0},\mathbf s^t).
\end{aligned}
\label{eq:mols_half_sensitivity}
\end{equation}
Under Assumption~\ref{ass:mols_moments}, the fixed multipliers satisfy
$\mathcal K_t(\widehat{\mathbf a})\leq\exp(\Delta^t)H^t$.
\end{molstheorem}

Theorem~\ref{thm:mols_one_shot} shows that common multiplicative
changes in SNR leave the population-based relative allocation unchanged.
For one-shot calibration, the relevant dynamics are therefore changes
in score ratios rather than changes in the absolute SNR level.
The bound separates calibration error from this relative drift.
The inverse-square-root map also halves relative score errors on
the logarithmic scale.

\paragraph{Finite-sample estimation}
Under local i.i.d. mini-batch sampling with finite fourth moments,
Corollary~\ref{cor:mols_finite_sample} gives an $N^{-1/2}$
calibration-error bound at fixed confidence and a fixed positive
SNR stabilizer. The estimates approach the corresponding stabilized
population scores.
Table~\ref{tab:snr_samples_ppl} complements this result with an
empirical study of calibration cost and final model quality.

\begin{figure*}[!t]
    \centering
    \begin{subfigure}[b]{0.32\linewidth}
        \centering
        \includegraphics[width=\linewidth]{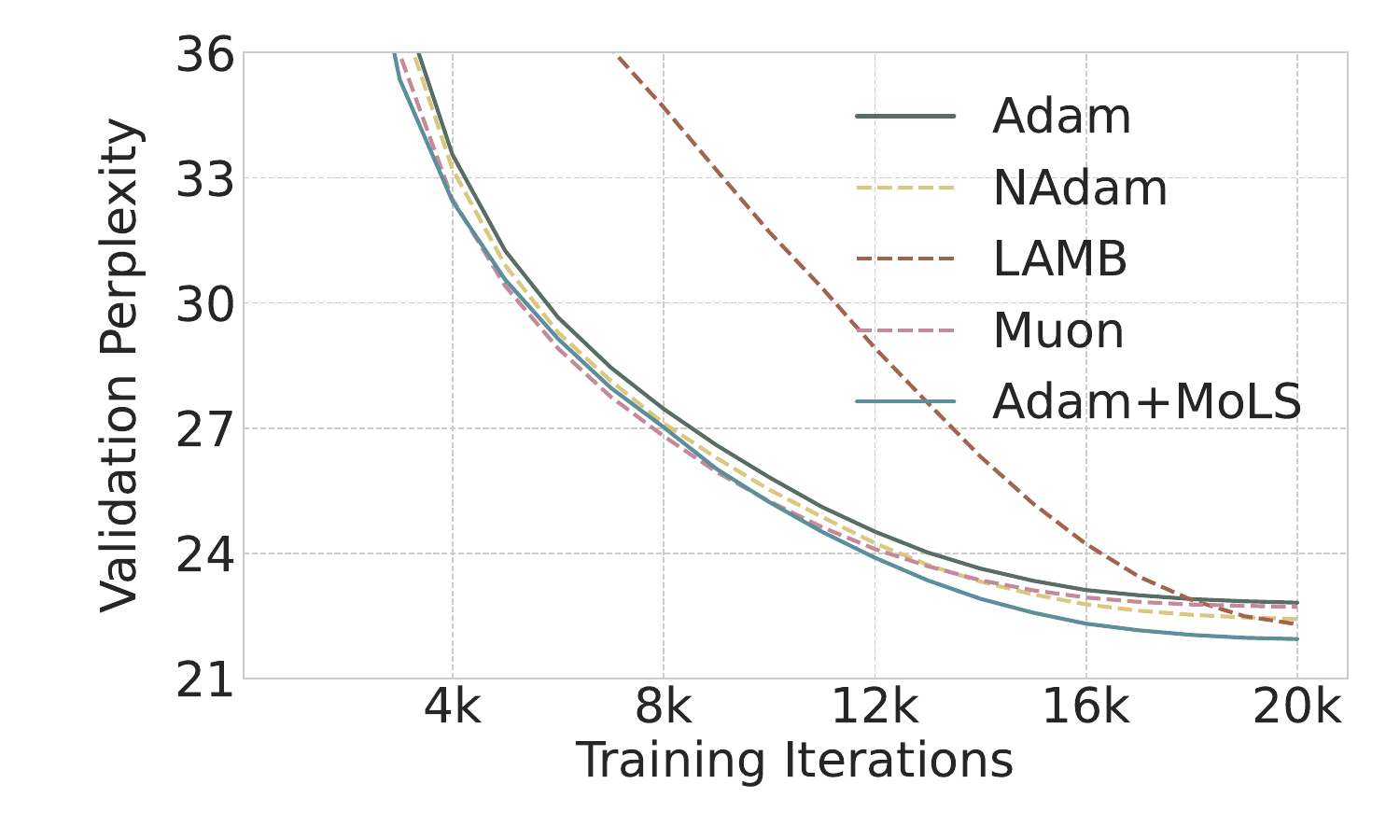}
        \caption{LLaMA-130M}
    \end{subfigure}\hfill
    \begin{subfigure}[b]{0.32\linewidth}
        \centering
        \includegraphics[width=\linewidth]{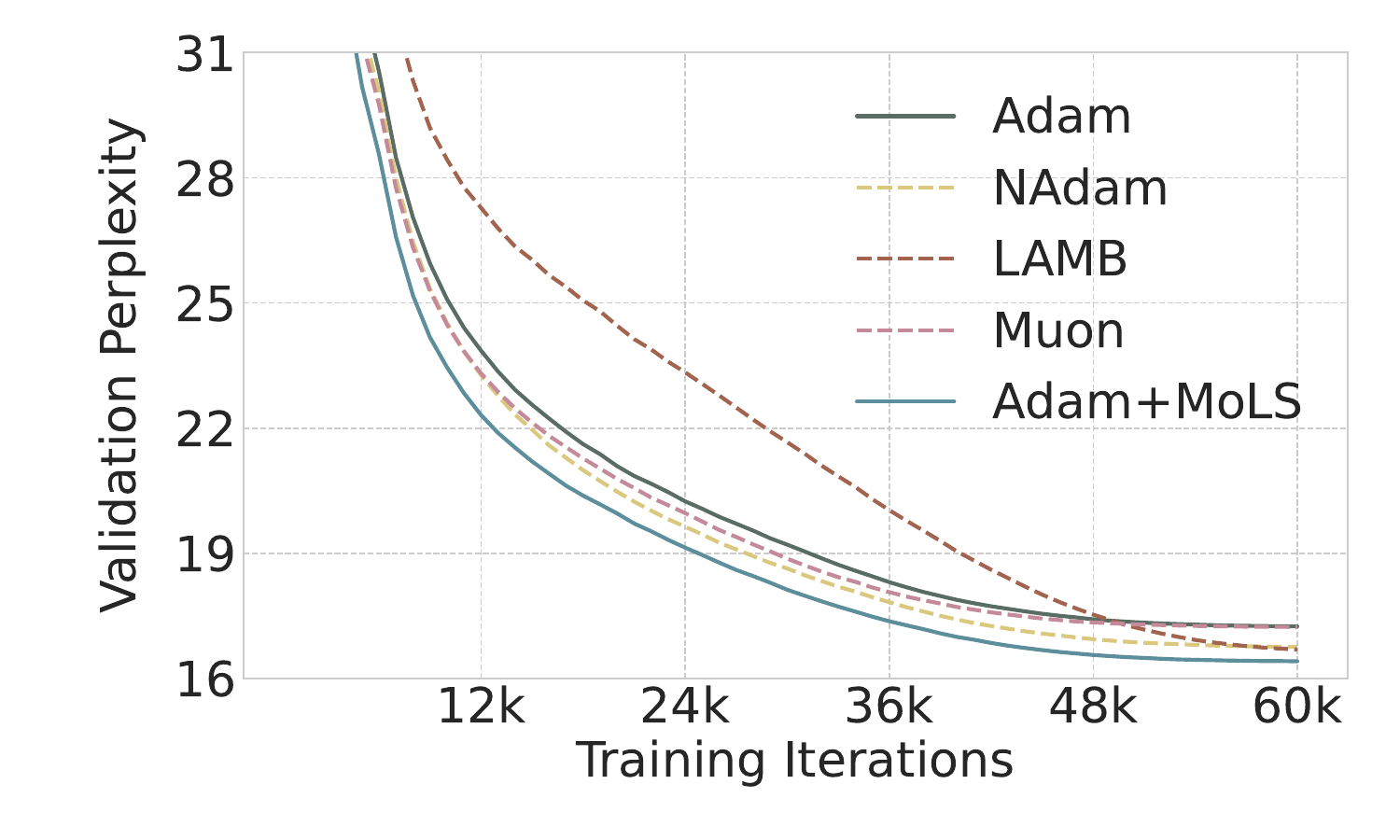}
        \caption{LLaMA-350M}
    \end{subfigure}\hfill
    \begin{subfigure}[b]{0.32\linewidth}
        \centering
        \includegraphics[width=\linewidth]{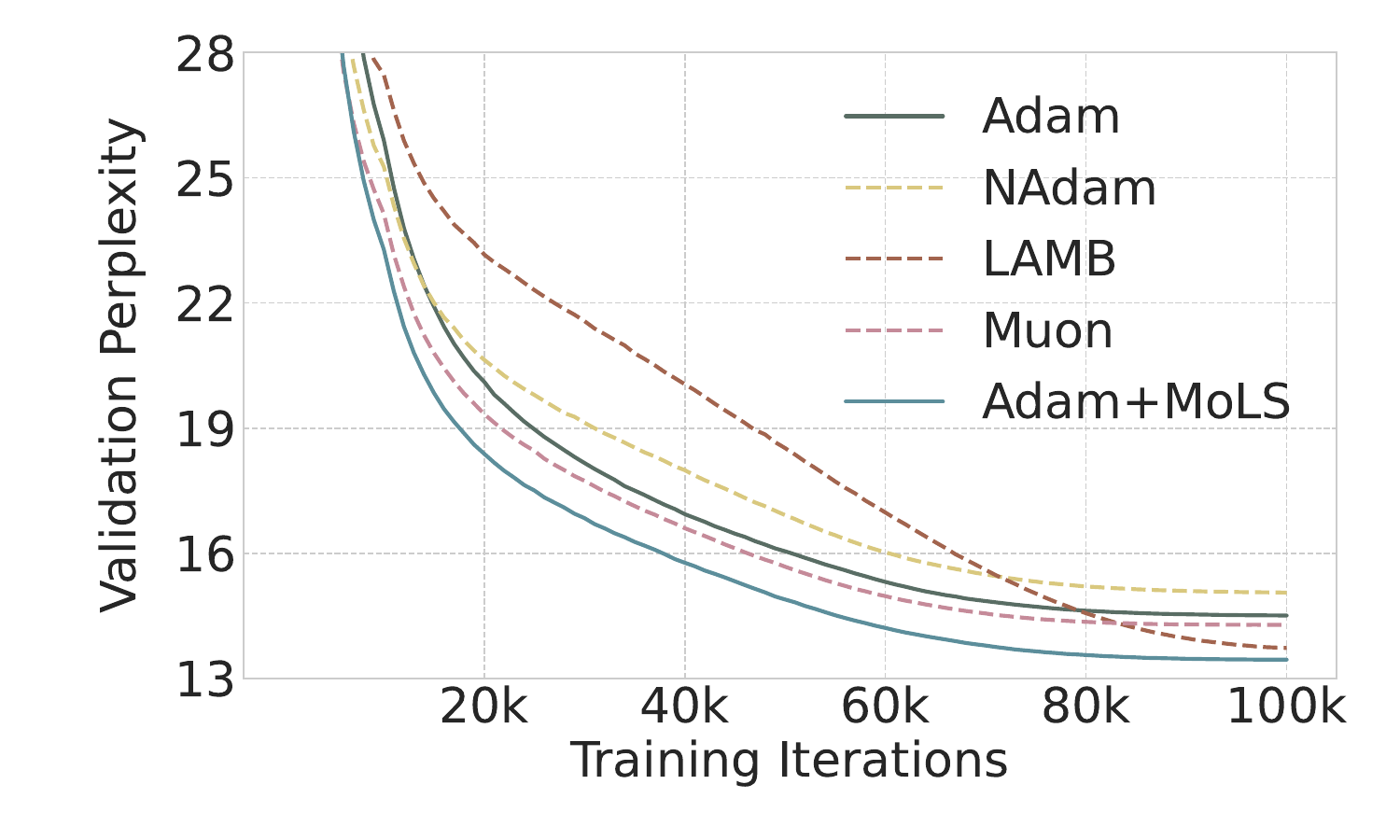}
        \caption{LLaMA-1.3B}
    \end{subfigure}
    \caption{Validation PPL curves for pre-training LLaMA-130M to 1.3B on C4.}
    \label{fig:val_ppl_llama_130m_1b_c4}
\end{figure*}

\begin{table*}[!t]
    \centering
    \resizebox{.75\linewidth}{!}{%
    \begin{tabular}{l|cccc}
    \toprule
    Methods & \textbf{60M} & \textbf{130M} & \textbf{350M} & \textbf{1.3B} \\
    \midrule
    \multicolumn{5}{l}{\gray{\textit{Pre-training with Full-Rank Optimizers}}} \\
    \grow Adam & $29.55_{\pm .02}$ & $22.82_{\pm .00}$ & $17.25_{\pm .00}$ & $14.51_{\pm .00}$ \\
    NAdam & $28.95_{\pm .04}$ & $22.42_{\pm .02}$ & $16.76_{\pm .00}$ & $15.06_{\pm .00}$ \\
    LAMB & $\mathbf{28.39}_{\pm .11}$ & $22.30_{\pm .05}$ & $16.69_{\pm .05}$ & $13.74_{\pm .00}$ \\
    Muon & $28.99_{\pm .03}$ & $22.73_{\pm .01}$ & $17.23_{\pm .01}$ & $14.29_{\pm .00}$ \\
    SGG & $30.31_{\pm .16}$ & $22.18_{\pm .08}$ & $17.28_{\pm .02}$ & $14.30_{\pm .00}$ \\
    SOAP & $29.02_{\pm .27}$ & $22.59_{\pm .14}$ & $17.03_{\pm .12}$ & $14.62_{\pm .00}$ \\
    \brow Adam+\name & $28.48_{\pm .10}$ & $\mathbf{21.94}_{\pm .06}$ & $\mathbf{16.41}_{\pm .03}$ & $\mathbf{13.45}_{\pm .00}$ \\
    {\gray{$\Delta$ PPL vs. Adam}} & \gbf{-1.07} & \gbf{-0.88} & \gbf{-0.84} & \gbf{-1.06} \\
    \midrule
    \multicolumn{5}{l}{\gray{\textit{Pre-training with Memory-efficient Optimizers}}} \\
    \grow Adam-mini & $29.63_{\pm .05}$ & $23.73_{\pm .01}$ & $17.83_{\pm .01}$ & $15.10_{\pm .00}$ \\
    GaLore & $33.24_{\pm .03}$ & $25.22_{\pm .02}$ & $18.67_{\pm .01}$ & $14.90_{\pm .00}$ \\
    APOLLO & $29.92_{\pm .15}$ & $22.86_{\pm .20}$ & $\mathbf{16.66}_{\pm .07}$ & $14.20_{\pm .00}$ \\
    Lion & $32.46_{\pm .08}$ & $25.47_{\pm .09}$ & $21.25_{\pm .04}$ & $15.69_{\pm .00}$ \\
    Adafactor & $32.57_{\pm .41}$ & $23.98_{\pm .22}$ & $17.74_{\pm .13}$ & $15.19_{\pm .00}$ \\
    \brow Adam-mini+\name & $\mathbf{28.49}_{\pm .07}$ & $\mathbf{21.91}_{\pm .07}$ & $16.86_{\pm .03}$ & $\mathbf{13.57}_{\pm .00}$ \\
    {\gray{$\Delta$ PPL vs. Adam-mini}} & \gbf{-1.14} & \gbf{-1.82} & \gbf{-0.97} & \gbf{-1.53} \\
    \midrule
    Training tokens & 1.3B & 2.6B & 7.8B & 13.1B \\
    \bottomrule
    \end{tabular}%
    }
    \caption{\textbf{Final validation PPL for LLaMA pre-training on C4 (lower is better).} Subscripts give the available three-run variability estimates for 60M--350M. Bold marks the lowest PPL within each optimizer group and model size. Baseline and \name{} rows are shaded (\sethlcolor{lightblue}\hl{blue background} identifies the proposed method). \gbf{Green} values give absolute PPL reductions from the corresponding base optimizer.}\label{tab:comp_c4_pt}
\end{table*}

\section{Experiments}\label{sec:exp}
We evaluate \name{} on language-model pre-training and downstream
fine-tuning. We report predictive performance, training and validation
trajectories, and end-to-end training time.
We also study module SNR patterns and calibration design choices.
Appendix~\ref{app:experimental_details} provides implementation details.

\subsection{Experimental Setup}\label{sec:exp_setup}
\paragraph{Pre-training configurations}
We follow the LLaMA configurations of Zhao et al.~\cite{zhao2024galore}
and train models from 60M to 7B parameters.
The default sequence length is 256 and total batch size is 512.
We use BF16 precision and gradient clipping at 1.0.
The base learning rate warms up linearly over the first 10\% of
training, then decays with a cosine schedule to 10\% of its initial
value.
For \name{}, calibration takes place after approximately 1\% of
the total training steps. The module multipliers are then introduced
over the remaining warmup, from about 1\% to 10\% of training,
without extending the schedule.
We train on C4~\cite{raffel2020exploring_c4} and
OpenWebText~\cite{Gokaslan2019OpenWeb}, using gradient checkpointing
for the 3B and 7B models.
Architecture specifications and selected learning rates appear in
Appendix~\ref{app:experimental_details}.

\paragraph{Baselines and hyperparameter selection}
Our full-parameter baselines are
Adam~\cite{kingma2014adam,loshchilov2017decoupled_adamw},
NAdam~\cite{dozat2016incorporating_nadamw},
LAMB~\cite{You2020Large_lamb},
Muon~\cite{jordan2024muon,liu2025muon},
SGG~\cite{li2025taming_sgg}, and SOAP~\cite{vyas2024soap}.
Memory-efficient baselines include Adam-mini~\cite{zhang2025adammini},
GaLore~\cite{zhao2024galore},
APOLLO~\cite{zhu2024apollosgdlikememoryadamwlevel},
Lion~\cite{chen2023symbolic_optimization}, and
Adafactor~\cite{shazeer2018adafactor}.
SGG results are taken from the cited work. All other results come
from our experiments.
For our pre-training runs, we search learning rates over
$\{1\mathrm{e}{-4},2.5\mathrm{e}{-4},5\mathrm{e}{-4},1\mathrm{e}{-3},2.5\mathrm{e}{-3},5\mathrm{e}{-3},1\mathrm{e}{-2},2.5\mathrm{e}{-2}\}$.
Method-specific parameters are given in the Appendix.
We compare final validation perplexity (PPL), learning curves,
and training time.

\subsection{Pre-Training LLaMA Models}\label{sec:exp_pretrain}
\begin{figure*}[!t]
    \centering
    \begin{subfigure}[b]{0.32\linewidth}
        \centering
        \includegraphics[width=\linewidth]{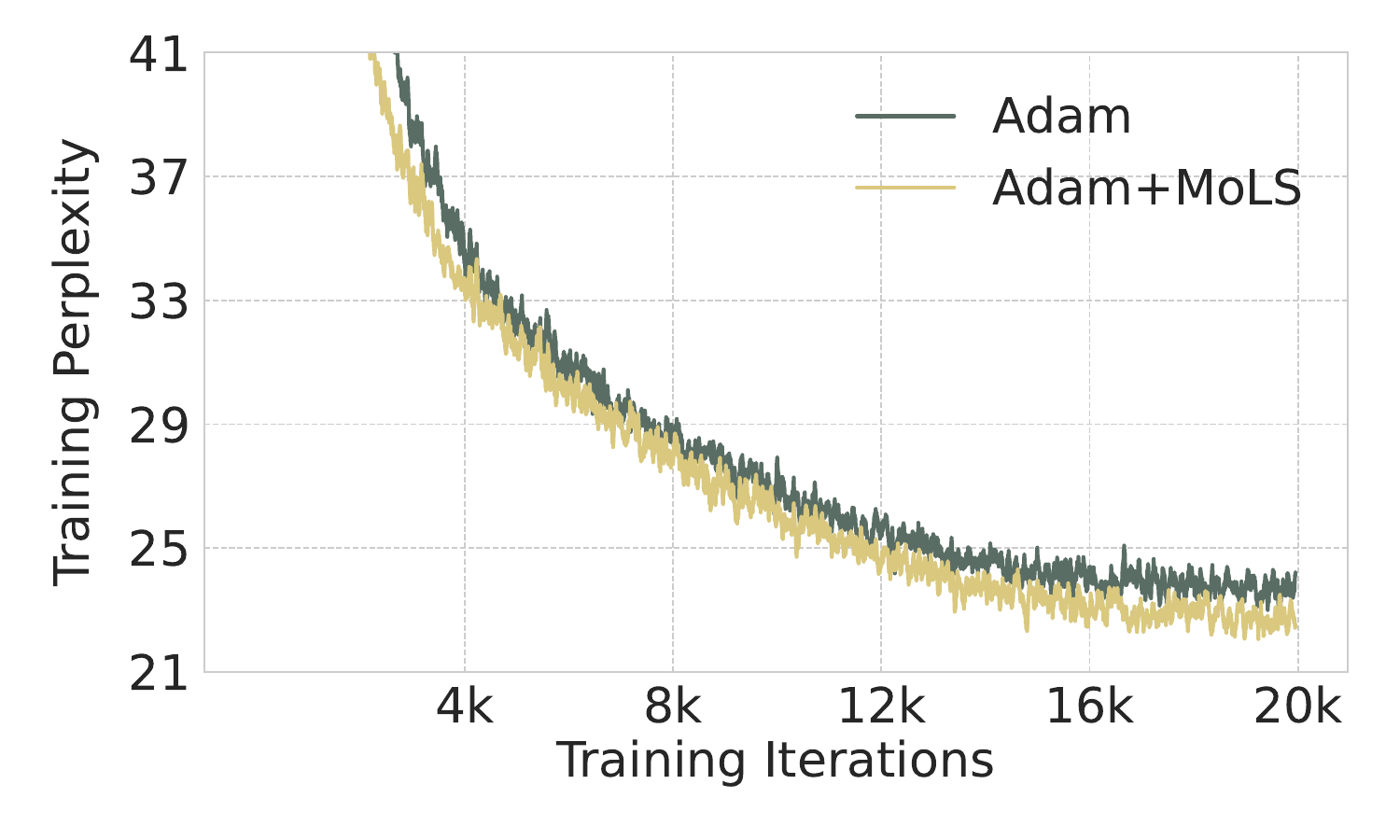}
        \caption{LLaMA-130M}
    \end{subfigure}\hfill
    \begin{subfigure}[b]{0.32\linewidth}
        \centering
        \includegraphics[width=\linewidth]{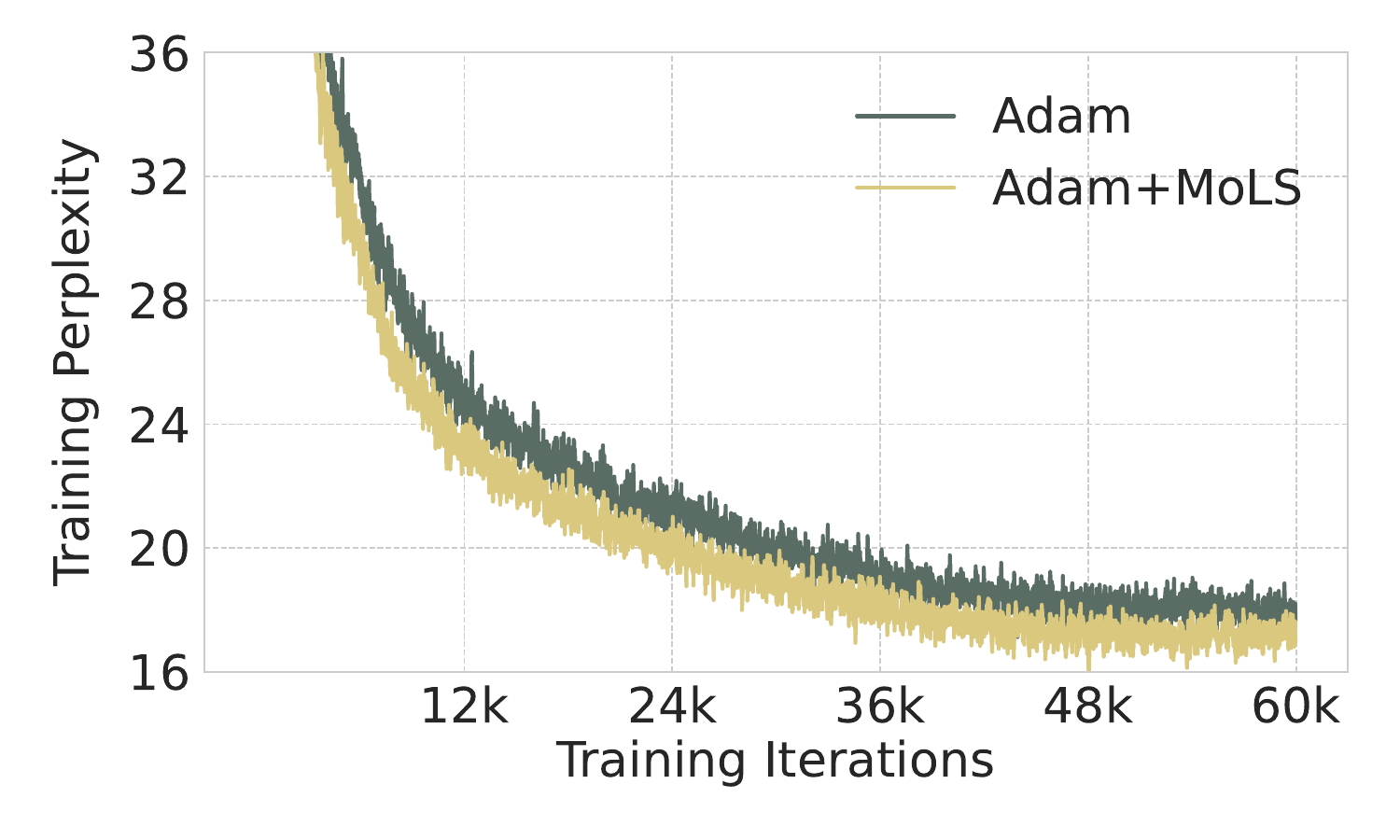}
        \caption{LLaMA-350M}
    \end{subfigure}\hfill
    \begin{subfigure}[b]{0.32\linewidth}
        \centering
        \includegraphics[width=\linewidth]{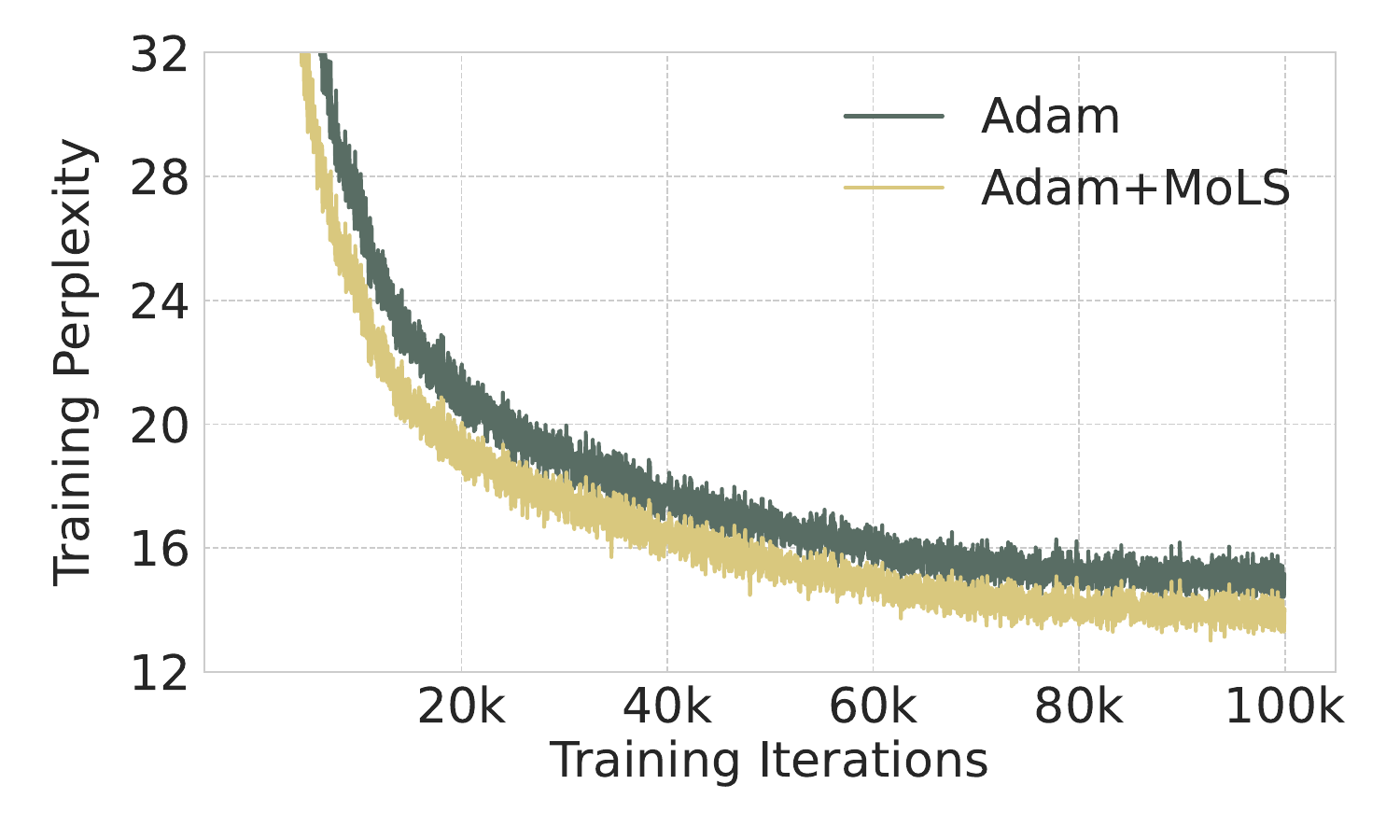}
        \caption{LLaMA-1.3B}
    \end{subfigure}
    \caption{Training PPL curves for pre-training LLaMA-130M to 1.3B on C4, smoothed over 50 iterations.}
    \label{fig:train_ppl_llama_130m_1b_c4}
\end{figure*}

\paragraph{Performance on C4}
Table~\ref{tab:comp_c4_pt} compares LLaMA models from 60M to 1.3B
parameters. Adding \name{} to Adam reduces final validation PPL
by 1.07, 0.88, 0.84, and 1.06 at the four scales.
The reductions for Adam-mini are 1.14, 1.82, 0.97, and 1.53.
Thus, \name{} improves both base optimizers at all four model sizes.
Adam+\name{} leads the full-parameter group at 130M, 350M, and 1.3B.
Adam-mini+\name{} leads the memory-efficient group at 60M, 130M,
and 1.3B.
LAMB gives the lowest full-parameter PPL at 60M, and APOLLO leads
the memory-efficient group at 350M.
Figures~\ref{fig:val_ppl_llama_130m_1b_c4} and
\ref{fig:train_ppl_llama_130m_1b_c4} show the corresponding validation
and training trajectories.

\paragraph{Training dynamics}
Figure~\ref{fig:train_ppl_llama_130m_1b_c4} compares Adam and
Adam+\name{} on LLaMA-130M, 350M, and 1.3B.
The training curves, smoothed over 50 iterations, show lower PPL
with \name{} at matched iteration counts.
The validation curves in Figure~\ref{fig:val_ppl_llama_130m_1b_c4}
show corresponding gains in held-out perplexity.

\begin{figure*}[!t]
    \centering
    \begin{subfigure}[b]{0.32\linewidth}
        \centering
        \includegraphics[width=\linewidth]{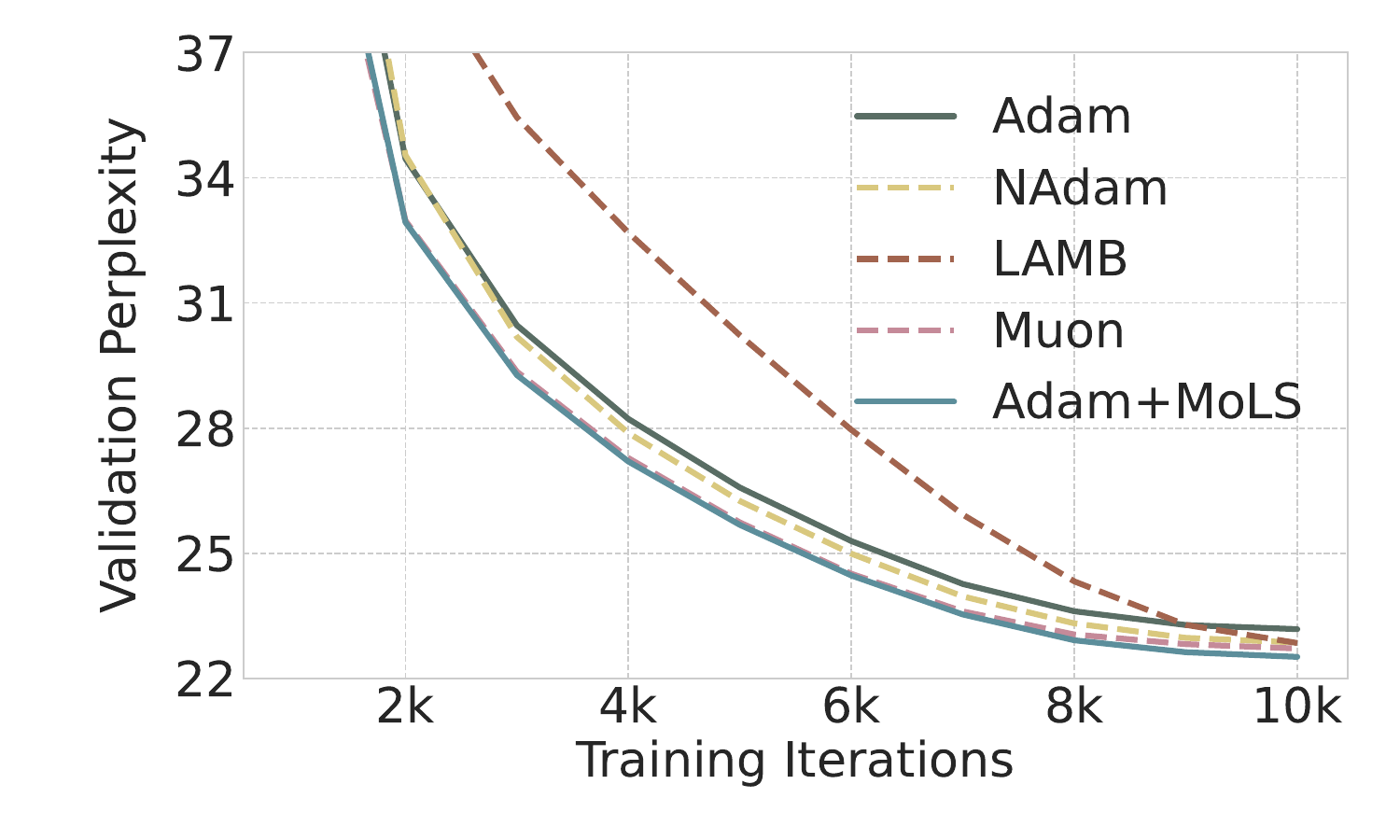}
        \caption{LLaMA-60M}
    \end{subfigure}\hfill
    \begin{subfigure}[b]{0.32\linewidth}
        \centering
        \includegraphics[width=\linewidth]{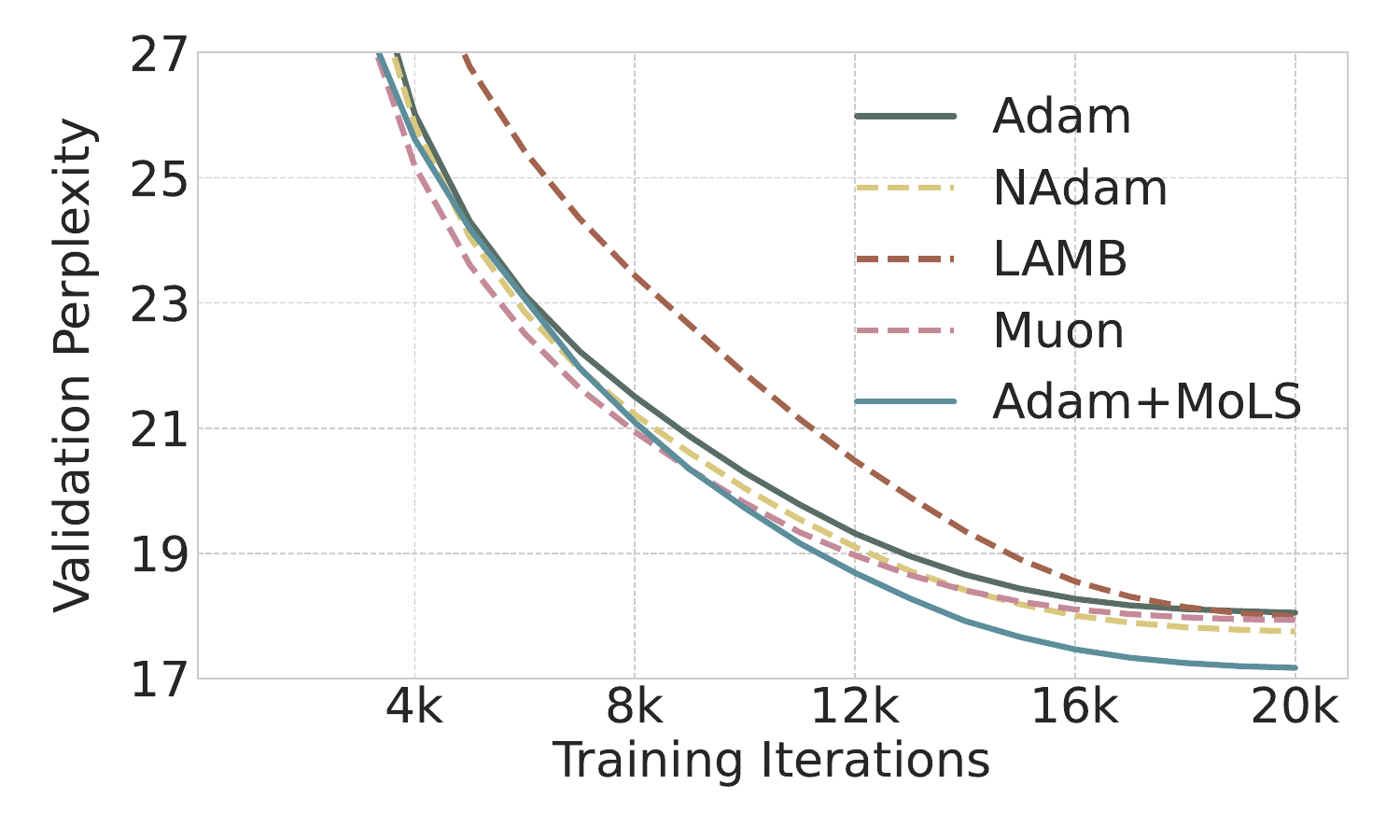}
        \caption{LLaMA-130M}
    \end{subfigure}\hfill
    \begin{subfigure}[b]{0.32\linewidth}
        \centering
        \includegraphics[width=\linewidth]{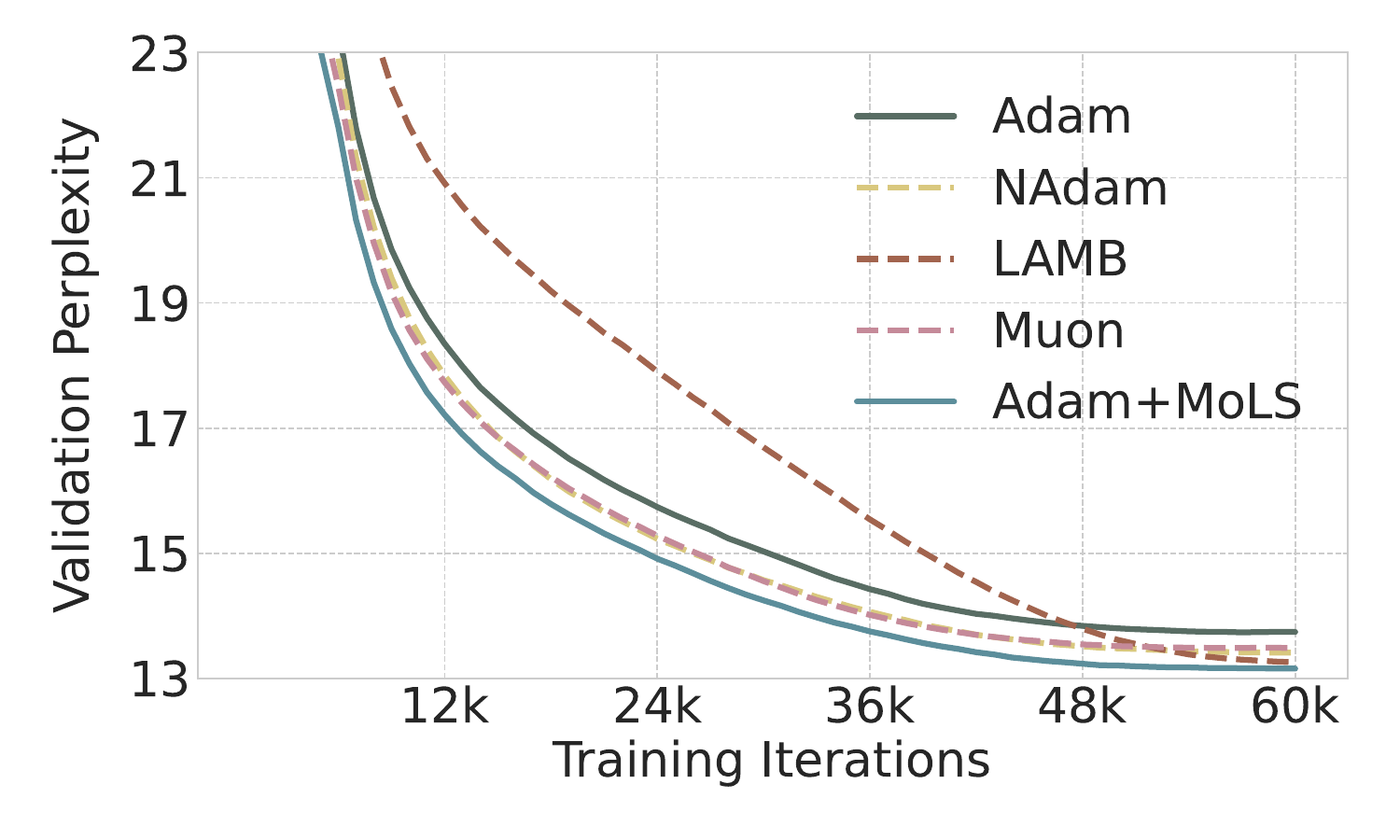}
        \caption{LLaMA-350M}
    \end{subfigure}
    \caption{PPL curves for pre-training LLaMA-60M to 350M on OpenWebText.}
    \label{fig:openwebtext_ppl}
\end{figure*}

\paragraph{Performance on OpenWebText}
We also pre-train LLaMA-60M, 130M, and 350M on OpenWebText.
Figure~\ref{fig:openwebtext_ppl} shows lower validation PPL for
Adam+\name{} than Adam at matched training steps.
These results extend the pre-training gains to a second dataset.

\paragraph{Scaling to 3B and 7B parameters}
We combine \name{} with 8-bit Adam~\cite{dettmers20218bit} to
pre-train LLaMA-3B and 7B with gradient checkpointing.
Table~\ref{tab:c4_3b_7b_PPL} compares 8-bit Adam, Muon, GaLore,
and APOLLO at matched training steps.
8-bit Adam+\name{} gives the lowest PPL at each checkpoint in the
table for both model sizes.
It reaches 13.07 PPL on 3B at 120K steps and 11.54 on 7B at
150K steps. The corresponding 8-bit Adam results are 14.19 and
13.23, giving absolute improvements of 1.12 and 1.69.

\begin{table}[!t]
    \centering
    \setlength{\tabcolsep}{1.4mm}
    \resizebox{\linewidth}{!}{%
    \begin{tabular}{l|ccccc}
    \toprule
    Methods & \textbf{30K} & \textbf{60K} & \textbf{90K} & \textbf{120K} & \textbf{150K} \\
    \midrule
    \multicolumn{6}{l}{\gray{\textit{Pre-training LLaMA-3B}}} \\
    \grow 8-bit Adam & 18.63 & 15.69 & 14.36 & 14.19 & -- \\
    Muon & 18.14 & 15.40 & 14.13 & 14.02 & -- \\
    GaLore & 18.44 & 15.98 & 14.90 & 14.73 & -- \\
    APOLLO & 19.49 & 15.40 & 14.09 & 13.75 & -- \\
    \brow 8-bit Adam+\name & \textbf{17.81} & \textbf{14.51} & \textbf{13.26} & \textbf{13.07} & -- \\
    \midrule
    \multicolumn{6}{l}{\gray{\textit{Pre-training LLaMA-7B}}} \\
    \grow 8-bit Adam & 18.73 & 16.13 & 14.56 & 13.40 & 13.23 \\
    Muon & 18.16 & 15.71 & 13.87 & 13.23 & 13.19 \\
    GaLore & 18.35 & 15.63 & 14.33 & 13.77 & 13.69 \\
    APOLLO & 18.49 & 15.17 & 13.54 & 12.80 & 12.63 \\
    \brow 8-bit Adam+\name & \textbf{16.22} & \textbf{13.77} & \textbf{12.34} & \textbf{11.70} & \textbf{11.54} \\
    \bottomrule
    \end{tabular}%
    }
    \caption{\textbf{Validation PPL across pre-training steps on C4 for LLaMA-3B and 7B with gradient checkpointing (lower is better).} Bold marks the best reported result at each checkpoint.}
    \label{tab:c4_3b_7b_PPL}
\end{table}

\paragraph{Generalization to GPT and Qwen architectures}
To assess transfer beyond LLaMA, we pre-train
GPT-2~\cite{radford2019languagegpt2} and
Qwen2.5~\cite{Qwen2024Qwen25} models on C4.
Table~\ref{tab:comp_c4_gpt_qwen} shows gains over Adam in all three
configurations. PPL falls from 31.09 to 29.65 on GPT2-125M,
from 22.19 to 21.66 on GPT2-355M, and from 16.87 to 16.34 on
Qwen2.5-350M.
Adam+\name{} gives the best result on GPT2-125M and Qwen2.5-350M.
NAdam leads on GPT2-355M with 21.55 PPL, compared with 21.66 for
Adam+\name{}.
Together with the LLaMA results, these gains demonstrate the
strategy's effectiveness across three model families.

\begin{table}[!t]
    \centering
    \resizebox{0.95\linewidth}{!}{%
    \begin{tabular}{l|ccc}
    \toprule
    Methods & \textbf{GPT2-125M} & \textbf{GPT2-355M} & \textbf{Qwen2.5-350M} \\
    \midrule
    \grow Adam & 31.09 & 22.19 & 16.87 \\
    NAdam & 30.26 & \textbf{21.55} & 16.48 \\
    LAMB & 30.50 & 21.70 & 16.72 \\
    Muon & 30.26 & 22.39 & 16.78 \\
    \brow Adam+\name & \textbf{29.65} & 21.66 & \textbf{16.34} \\
    \midrule
    Training tokens & 2.6B & 7.8B & 7.8B \\
    \bottomrule
    \end{tabular}%
    }
    \caption{Final validation PPL for GPT-2 and Qwen2.5 pre-training on C4 (lower is better). Bold indicates the best method per model.}
    \label{tab:comp_c4_gpt_qwen}
\end{table}

\subsection{Efficient Fine-Tuning}\label{sec:fine_tune}
\paragraph{Setup and baselines}
We evaluate LLaMA-3.2-1B and 3B~\cite{grattafiori2024llama3},
Gemma3-1B~\cite{team2025gemma3}, and
Qwen2.5-7B~\cite{Qwen2024Qwen25}.
Supervised fine-tuning (SFT) uses eight common-sense reasoning
benchmarks: Winogrande (WG)~\cite{sakaguchi2021winogrande},
PIQA~\cite{bisk2020piqa}, SIQA~\cite{sap2019socialiqa_siqa},
OpenBookQA (OBQA)~\cite{mihaylov2018can_openbookqa},
HellaSwag (HS)~\cite{zellers2019hellaswag},
BoolQ~\cite{clark2019boolq}, and
ARC-Easy/Challenge~\cite{clark2018thinkarc}.
We follow the evaluation protocol of prior
work~\cite{liu2024dora,zhu2024apollosgdlikememoryadamwlevel}.
The full-parameter baselines are Adam, NAdam, LAMB, and Muon.
For parameter-efficient fine-tuning (PEFT), we follow
LLaMA-Factory~\cite{zheng2024llamafactory} and evaluate on
MMLU~\cite{hendrycks2020measuring_mmlu}.
We compare LoRA~\cite{hu2022lora}, full-parameter fine-tuning,
GaLore, APOLLO, and LoRA+\name{}.
Appendix~\ref{app:experimental_details} provides hyperparameters.

\paragraph{Common-sense reasoning}
Table~\ref{tab:common_sense_sft} reports SFT results for
LLaMA-3.2-1B.
Adam+\name{} achieves the highest average accuracy at 69.18\%,
compared with 68.32\% for Adam and 68.55\% for Muon.
It improves Winogrande accuracy from 68.19\% to 70.47\% and
PIQA accuracy from 76.12\% to 77.61\%.
These results show the benefit of module-wise calibration for
supervised downstream training.

\begin{table*}[!t]
    \centering
    \resizebox{0.95\linewidth}{!}{%
    \begin{tabular}{l|cccccccc|c}
    \toprule
    Method & \textbf{WG} & \textbf{PIQA} & \textbf{SIQA} & \textbf{OBQA} & \textbf{HS} & \textbf{BoolQ} & \textbf{ARC-E} & \textbf{ARC-C} & \textbf{Avg.} \\
    \midrule
    \grow Adam & 68.19 & 76.12 & 72.36 & 69.00 & 69.19 & 64.34 & 72.22 & 55.12 & 68.32 \\
    NAdam & 67.91 & 76.20 & 70.77 & 68.21 & 69.03 & 63.77 & 73.31 & 55.20 & 68.05 \\
    LAMB & 68.33 & 74.81 & 71.56 & 69.07 & 68.21 & 64.00 & 71.48 & 54.92 & 67.79 \\
    Muon & 70.21 & 76.34 & 71.90 & 68.89 & 69.31 & 64.51 & 72.15 & 55.14 & 68.55 \\
    \brow Adam+\name & 70.47 & 77.61 & 72.41 & 70.33 & 69.18 & 64.40 & 73.73 & 55.36 & \textbf{69.18} \\
    \bottomrule
    \end{tabular}%
    }
    \caption{Accuracy (\%) on eight common-sense reasoning tasks after full-parameter SFT of LLaMA-3.2-1B. The best average is in bold.}
    \label{tab:common_sense_sft}
\end{table*}

\paragraph{Parameter-efficient fine-tuning}
Table~\ref{tab:mmlu_peft} reports MMLU accuracy using the learning-rate
grid in the caption.
LoRA+\name{} improves average accuracy over LoRA by 0.83, 0.76,
and 0.60 percentage points on Gemma3-1B, LLaMA-3.2-3B, and
Qwen2.5-7B, respectively.
It gives the highest average on LLaMA-3.2-3B and Qwen2.5-7B.
On Gemma3-1B, it reaches 26.31\%, close to APOLLO's 26.38\%.
The gains across all three models demonstrate that SNR-based
calibration complements LoRA.

\begin{table}[!t]
    \centering
    \resizebox{\linewidth}{!}{%
    \begin{tabular}{ll|ccccc}
    \toprule
    Models & Methods & \textbf{STEM} & \textbf{Soc.} & \textbf{Hum.} & \textbf{Other} & \textbf{Avg.} \\
    \midrule
    \multirow{5}{*}{Gemma3-1B}
    & Full & 27.63 & 26.84 & 25.12 & 25.14 & 26.04 \\
    & \cellcolor{gray94}LoRA & \cellcolor{gray94}26.61 & \cellcolor{gray94}25.93 & \cellcolor{gray94}24.78 & \cellcolor{gray94}25.02 & \cellcolor{gray94}25.48 \\
    & GaLore & 27.40 & 26.71 & 24.80 & 25.69 & 25.99 \\
    & APOLLO & 27.47 & 27.14 & 25.61 & 25.76 & \textbf{26.38} \\
    & \cellcolor{lightblue}LoRA+\name & \cellcolor{lightblue}27.51 & \cellcolor{lightblue}26.96 & \cellcolor{lightblue}25.67 & \cellcolor{lightblue}25.49 & \cellcolor{lightblue}26.31 \\
    \midrule
    \multirow{5}{*}{LLaMA3.2-3B}
    & Full & 46.55 & 65.81 & 49.16 & 63.17 & 55.48 \\
    & \cellcolor{gray94}LoRA & \cellcolor{gray94}47.22 & \cellcolor{gray94}63.41 & \cellcolor{gray94}49.56 & \cellcolor{gray94}61.57 & \cellcolor{gray94}54.86 \\
    & GaLore & 46.62 & 65.58 & 48.37 & 63.33 & 55.22 \\
    & APOLLO & 46.16 & 65.91 & 49.03 & 62.80 & 55.29 \\
    & \cellcolor{lightblue}LoRA+\name & \cellcolor{lightblue}47.34 & \cellcolor{lightblue}65.77 & \cellcolor{lightblue}49.61 & \cellcolor{lightblue}62.41 & \cellcolor{lightblue}\textbf{55.62} \\
    \midrule
    \multirow{5}{*}{Qwen2.5-7B}
    & Full & \multicolumn{4}{c}{OOM} & -- \\
    & \cellcolor{gray94}LoRA & \cellcolor{gray94}67.69 & \cellcolor{gray94}82.91 & \cellcolor{gray94}66.29 & \cellcolor{gray94}75.72 & \cellcolor{gray94}72.41 \\
    & GaLore & 68.32 & 82.87 & 66.08 & 76.10 & 72.55 \\
    & APOLLO & 68.49 & 82.97 & 65.80 & 76.65 & 72.65 \\
    & \cellcolor{lightblue}LoRA+\name & \cellcolor{lightblue}68.71 & \cellcolor{lightblue}82.87 & \cellcolor{lightblue}66.41 & \cellcolor{lightblue}77.21 & \cellcolor{lightblue}\textbf{73.01} \\
    \bottomrule
    \end{tabular}%
    }
    \caption{\textbf{MMLU accuracy (\%) for fine-tuning with one A100 40GB GPU.} The reported values are the best over the learning-rate grid $\{1\mathrm{e}{-5},2.5\mathrm{e}{-5},5\mathrm{e}{-5},1\mathrm{e}{-4},1.5\mathrm{e}{-4},2\mathrm{e}{-4}\}$. Bold indicates the best average per model.}
    \label{tab:mmlu_peft}
\end{table}

\subsection{Computational Efficiency}\label{sec:exp_efficiency}
We assess the cost of one-shot calibration through final validation
PPL and end-to-end training time.
Table~\ref{tab:time_ppl_comp} shows improved final PPL with less than
2\% additional end-to-end training time on LLaMA-1.3B/C4.
Adam+\name{} reduces PPL from 14.51 to 13.45, while runtime
increases from 54.18 to 55.08 hours (+1.66\%).
Adam-mini+\name{} reduces PPL from 15.10 to 13.57, with runtime
increasing from 53.10 to 54.01 hours (+1.71\%).
All methods run on 16 NVIDIA RTX 3090 24GB GPUs with a per-GPU
batch size of 32.
The comparison measures final model quality and total runtime at
the prescribed training budget.
Appendix~\ref{app:snr_details} describes calibration.

\begin{table}[!t]
    \centering
    \resizebox{\linewidth}{!}{%
    \begin{tabular}{l|ll}
    \toprule
    Methods & Final PPL & Training time (h) \\
    \midrule
    \grow Adam & 14.51 & 54.18 \\
    NAdam & 15.06 (\rbf{+3.79\%}) & 55.30 (\rbf{+2.07\%}) \\
    LAMB & 13.74 (\gbf{+5.31\%}) & 57.41 (\rbf{+5.96\%}) \\
    Muon & 14.29 (\gbf{+1.52\%}) & 91.83 (\rbf{+69.49\%}) \\
    \brow Adam+\name & \textbf{13.45} (\gbf{+7.31\%}) & 55.08 (\rbf{+1.66\%}) \\
    \midrule
    \grow Adam-mini & 15.10 & 53.10 \\
    GaLore & 14.90 (\gbf{+1.32\%}) & 65.63 (\rbf{+23.60\%}) \\
    APOLLO & 14.20 (\gbf{+5.96\%}) & 53.71 (\rbf{+1.15\%}) \\
    \brow Adam-mini+\name & \textbf{13.57} (\gbf{+10.13\%}) & 54.01 (\rbf{+1.71\%}) \\
    \bottomrule
    \end{tabular}%
    }
    \caption{\textbf{Final PPL and end-to-end training time for LLaMA-1.3B on C4.} Percentages denote relative PPL \gbf{gains} or \rbf{degradations}, and \rbf{increases} in total training time, relative to Adam (upper block) or Adam-mini (lower block). Positive green PPL gains indicate lower perplexity. }
    \label{tab:time_ppl_comp}
\end{table}

\section{Empirical Analysis and Ablation Studies}\label{sec:exp_analysis}
We examine the evidence for functional grouping, the sampling cost
of calibration, and its ability to replace manual multiplier search.
Longer-context and extended-budget runs assess the gains under
changes in training conditions.
These studies connect the design of \name{} to the performance
comparisons in Section~\ref{sec:exp}.

\subsection{Understanding SNR-Based Calibration}\label{sec:exp_snr_analysis}
\paragraph{Cross-scale consistency of module-wise scaling}
Table~\ref{tab:alpha_value} shows a consistent pattern of
module-specific scaling from LLaMA-60M to 7B.
With VO normalized to one, embedding multipliers range from
7.91 to 13.3 and head multipliers from 3.96 to 10.2.
QK and MLP multipliers range from 1.20 to 1.76 and from 1.31 to 1.55,
respectively.
Across all six scales, calibration assigns larger multipliers to
embeddings and output heads than to attention and MLP components.
The values adjust to model size while preserving this functional
distinction.

\begin{table}[!t]
    \centering
    \setlength{\tabcolsep}{10pt}
    \renewcommand{\arraystretch}{1.1}
    \begin{tabular}{l|ccccc}
    \toprule
    Models & $\alpha_{\mathrm{VO}}$ & $\alpha_{\mathrm{QK}}$ & $\alpha_{\mathrm{MLP}}$ & $\alpha_{\mathrm{Emb}}$ & $\alpha_{\mathrm{Head}}$ \\
    \midrule
    60M & 1.00 & 1.20 & 1.31 & 12.2 & 10.2 \\
    130M & 1.00 & 1.43 & 1.41 & 11.8 & 5.94 \\
    350M & 1.00 & 1.56 & 1.38 & 13.3 & 6.70 \\
    1.3B & 1.00 & 1.76 & 1.55 & 12.8 & 9.70 \\
    3B & 1.00 & 1.60 & 1.40 & 8.66 & 3.96 \\
    7B & 1.00 & 1.51 & 1.39 & 7.91 & 5.77 \\
    \bottomrule
    \end{tabular}
    \caption{Estimated module-wise scaling factors $\alpha_m$ for pre-training LLaMA-60M to 7B on C4. All factors are normalized to the VO reference module.}
    \label{tab:alpha_value}
\end{table}

\begin{figure*}[!t]
    \centering
    \begin{subfigure}[b]{0.98\linewidth}
        \centering
        \includegraphics[width=\linewidth]{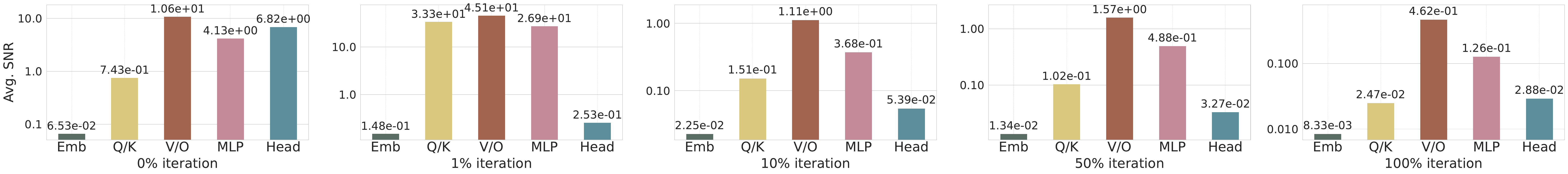}
        \caption{LLaMA-60M on OpenWebText.}
    \end{subfigure}
    \begin{subfigure}[b]{0.98\linewidth}
        \centering
        \includegraphics[width=\linewidth]{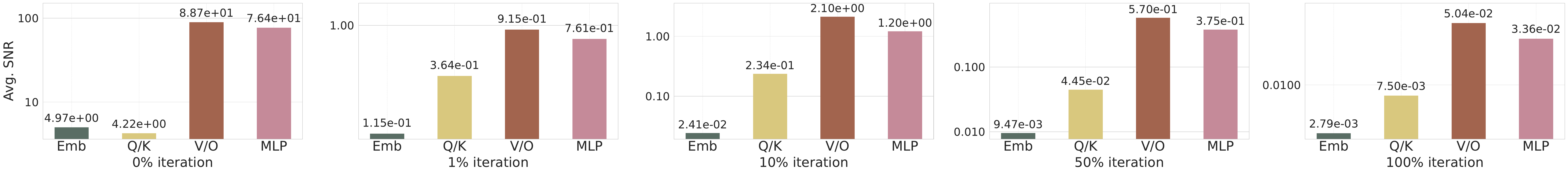}
        \caption{GPT2-125M on C4.}
    \end{subfigure}
    \caption{SNR variation across modules during pre-training under different model/data settings. GPT-2 uses a tied embedding/output head.}
    \label{fig:snr_gpt_open}
\end{figure*}

\paragraph{SNR dynamics across datasets and architectures}
Figure~\ref{fig:snr_gpt_open} shows SNR dynamics for LLaMA-60M on
OpenWebText and GPT2-125M on C4.
Both exhibit clear differences among functional groups, with scores
evolving through training.
GPT-2 ties its output head to the input embedding, so the shared
parameters are shown as a single group.
These measurements extend the evidence for module-wise heterogeneity
across datasets and architectures.
The configuration-dependent scores motivate calibration from the
current model's gradient statistics.

\subsection{Sensitivity and Design Choices}\label{sec:exp_ablation}
\paragraph{Sensitivity to the SNR estimation budget}
One-shot calibration is intended to use a short sampling stage.
We vary its budget to examine the trade-off between collection cost
and final model quality.
Table~\ref{tab:snr_samples_ppl} shows low sensitivity of final
validation PPL to the calibration budget on LLaMA-130M/C4.
Across budgets from 2,048 to 32,768 examples, PPL ranges from
21.93 to 22.04, while estimation time rises from 0.67 to 7.83 minutes.
The default budget of 16,384 examples gives 21.94 PPL with
3.73 minutes of estimation.
Using 2,048 examples gives 21.99 PPL with 0.67 minutes.
Smaller budgets therefore reduce calibration cost with little change
in final model quality.
Corollary~\ref{cor:mols_finite_sample} complements this performance
study with a consistency result for the stabilized estimates.

\begin{table}[!t]
    \centering
    \resizebox{0.95\linewidth}{!}{%
    \begin{tabular}{l|cc}
    \toprule
    Samples & Final PPL & SNR estimation time (min) \\
    \midrule
    2,048 & 21.99 & 0.67 \\
    4,096 & 22.04 & 1.16 \\
    8,192 & 21.96 & 2.02 \\
    16,384 & 21.94 & 3.73 \\
    32,768 & \textbf{21.93} & 7.83 \\
    \bottomrule
    \end{tabular}%
    }
    \caption{Sensitivity to the SNR estimation sample budget when pre-training LLaMA-130M on C4. Lower PPL is better.}
    \label{tab:snr_samples_ppl}
\end{table}

\begin{figure*}[!t]
    \centering
    \includegraphics[width=0.95\linewidth]{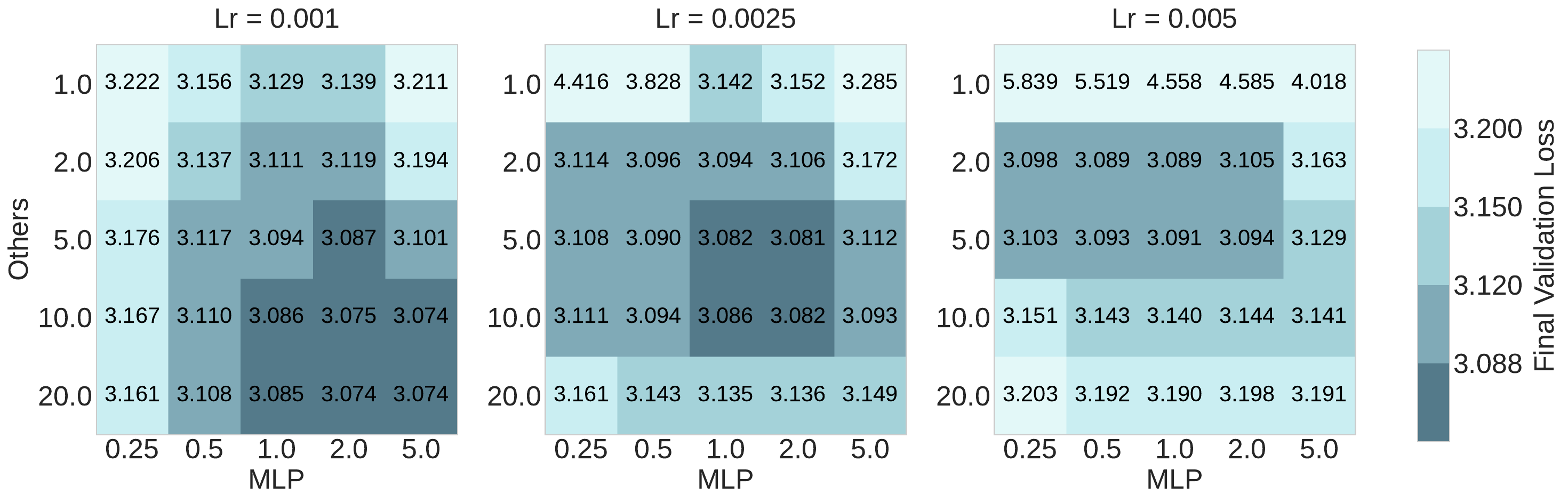}
    \caption{Final validation loss for LLaMA-130M pre-training on C4 under manually tuned module-wise learning rates. The search varies the base rate for QK/VO, the MLP multiplier, and the remaining-module multiplier. The best reported manual loss is 3.074, compared with 3.088 for \name.}
    \label{fig:llama_c4_hand_tune}
\end{figure*}

\paragraph{Comparison with hand-tuned module-wise learning rates}
We test whether early SNR estimates can replace a separate search
over module-specific multipliers.
For this comparison, we use a structured learning-rate search on
LLaMA-130M/C4.
The QK/VO base rate is chosen from
$\{1\mathrm{e}{-3},2.5\mathrm{e}{-3},5\mathrm{e}{-3}\}$.
The MLP multiplier uses
$\alpha_{\mathrm{MLP}}\in\{0.25,0.5,1.0,2.0,5.0\}$,
and the remaining modules use
$\alpha_{\mathrm{others}}\in\{1.0,2.0,5.0,10.0,20.0\}$.
Figure~\ref{fig:llama_c4_hand_tune} reports a loss of 3.088 for
\name{}, compared with 3.127 for vanilla Adam and 3.074 for the
best manually searched configuration.
\name{} reduces the baseline loss by 0.039 and comes within 0.014
of the best searched result, without a separate multiplier search.

\paragraph{Ablation on module grouping}
We progressively refine the parameter partition to test which
functional distinctions improve calibration.
Table~\ref{tab:ablation_module} reports the results on LLaMA-130M/C4.
The coarse Attn/MLP grouping gives a loss of 3.139, compared with
3.127 for vanilla Adam.
Separating embeddings reduces the loss to 3.103.
Splitting attention into QK and VO reduces it to 3.092, and
separating the head further reduces it to 3.088.
Each refinement improves on the preceding grouped configuration,
supporting the five-group partition used by \name{}.

\begin{table}[!t]
    \centering
    \begin{tabular}{l|cc}
    \toprule
    Grouping strategy & Loss & $\Delta$ vs. vanilla \\
    \midrule
    \grow Vanilla & 3.127 & -- \\
    \midrule
    Attn, MLP & 3.139 & \rbf{+0.012} \\
    Attn, MLP, Emb & 3.103 & \gbf{-0.024} \\
    QK, VO, MLP, Emb & 3.092 & \gbf{-0.035} \\
    \brow QK, VO, MLP, Emb, Head & \textbf{3.088} & \gbf{-0.039} \\
    \bottomrule
    \end{tabular}
    \caption{\textbf{Ablation of SNR estimation and learning-rate scaling groups for LLaMA-130M pre-training on C4.} \gbf{Green} and \rbf{red} denote improvements and degradations in validation loss relative to vanilla Adam, respectively.}
    \label{tab:ablation_module}
\end{table}

\subsection{Robustness across Training Regimes}\label{sec:exp_robustness}
\begin{figure}[!t]
    \centering
    \includegraphics[width=0.8\linewidth]{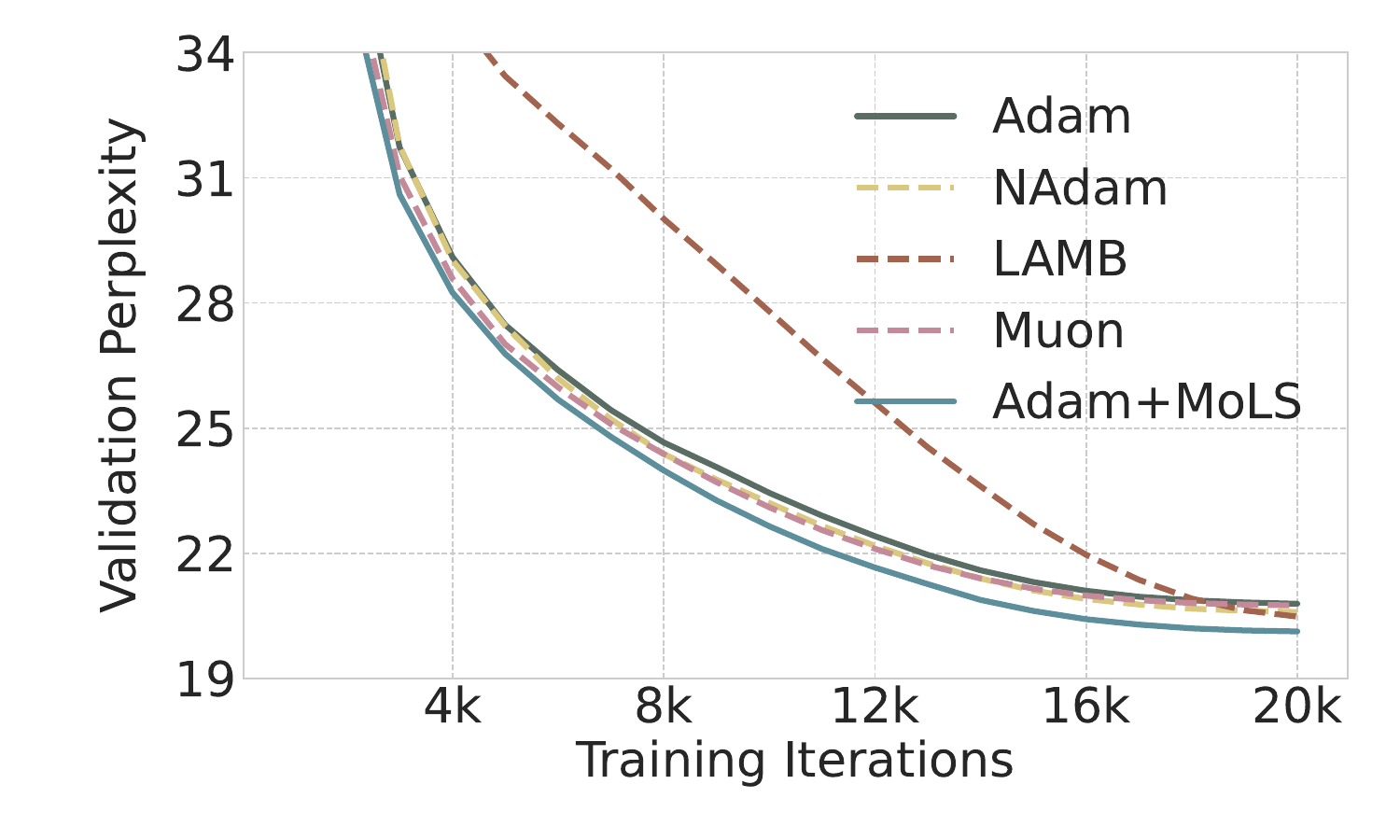}
    \caption{Validation PPL during LLaMA-130M pre-training with sequence length 1,024.}
    \label{fig:llama_130m_1024}
\end{figure}

\paragraph{Longer context lengths}
To test calibration with longer input sequences, we increase the
LLaMA-130M pre-training sequence length from 256 to 1,024.
Figure~\ref{fig:llama_130m_1024} shows lower validation PPL for
\name{} than the baselines at matched training steps.
The optimization gains therefore extend to this longer-context
configuration.

\begin{figure}[!t]
    \centering
    \includegraphics[width=0.8\linewidth]{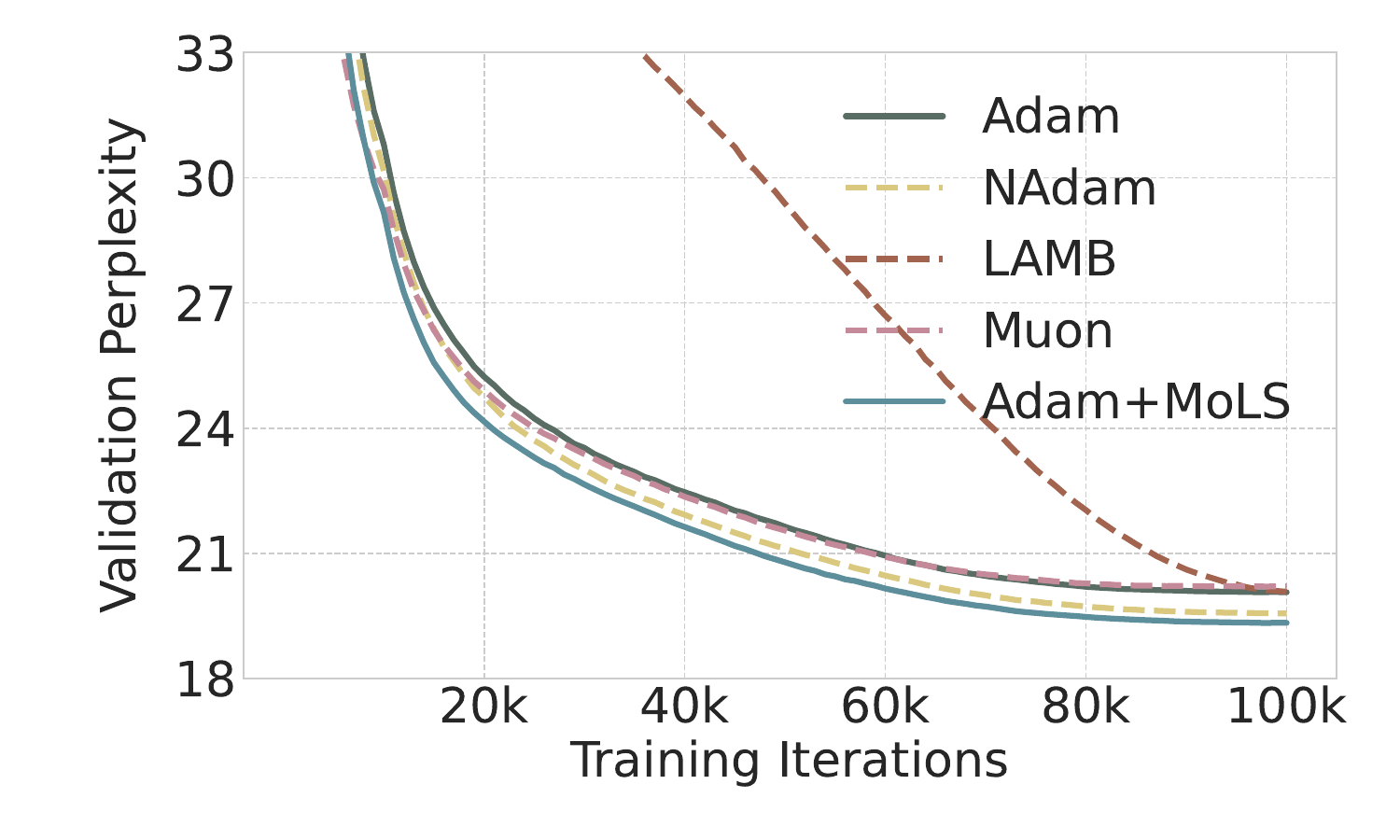}
    \caption{PPL during extended LLaMA-130M pre-training on C4 for 100K iterations (approximately five times Chinchilla-style compute-optimal tokens).}
    \label{fig:llama_long_train}
\end{figure}

\paragraph{Extended training token budgets}
We test whether the benefits of early calibration persist over a
longer training horizon.
We extend LLaMA-130M pre-training on C4 to 100K iterations,
approximately five times the Chinchilla-style compute-optimal token
budget for this model size~\cite{hoffmann2022scalinglaw}.
Figure~\ref{fig:llama_long_train} shows that \name{} maintains lower
validation PPL than the baselines over the extended schedule.
The advantage therefore persists beyond early training.

\section{Discussion and Limitations}
\label{sec:discussion}

\paragraph{Static versus dynamic calibration}
\name{} uses one-shot calibration to address persistent functional
differences while paying the statistics-collection cost once.
The global schedule controls the overall learning-rate evolution,
and the fixed multipliers determine the relative allocation.
Theorem~\ref{thm:mols_one_shot} shows that common multiplicative
SNR changes cancel from this allocation and quantifies the effect of relative drift.
The extended-training results in Figure~\ref{fig:llama_long_train}
show that the benefits of early calibration persist over a longer
schedule.
Low-frequency or event-triggered re-estimation is a natural extension
when relative module statistics change substantially.

\paragraph{Estimation robustness and module grouping}
The budget study in Table~\ref{tab:snr_samples_ppl} shows little
variation in final PPL on LLaMA-130M/C4.
Small-batch training and distribution shifts are useful settings
for studying robust or uncertainty-aware calibration.
\name{} currently uses predefined functional groups, with the
five-group partition giving the best result in
Table~\ref{tab:ablation_module}.
Automatic grouping and finer layer-wise calibration offer ways to
capture additional heterogeneity, at the cost of more detailed
statistics.

\paragraph{Reference-module normalization}
VO is the reference in the LLaMA experiments.
For fixed SNR scores, changing the reference rescales all multipliers
by a common factor and preserves their pairwise ratios.
Reference selection therefore controls the overall scale, while
measured SNR ratios determine the relative allocation.
Automatic reference selection or reference-free normalization can be
studied together with the global learning-rate configuration.

\paragraph{Applicability and future extensions}
\name{} operates through parameter-group learning rates and
integrates with Adam, Adam-mini, 8-bit Adam, and LoRA fine-tuning.
The rule accepts a chosen partition of trainable parameters, with
five functional groups providing the configuration studied here.
Extensions to mixture-of-experts, non-Transformer, and multimodal
models can use groups matched to their structure.
Future work will examine these groupings and the interaction of
SNR-based calibration with other optimizer update rules.

\section{Conclusion}
We presented \textbf{Module-wise Learning Rate Scaling via SNR
(\name)}, a lightweight method that converts early gradient statistics
into module-specific learning rates without a separate multiplier
search. It uses the existing warmup interval to introduce the
module multipliers, preserving the original training schedule.
Our analysis identifies structural sources of SNR
heterogeneity and quantifies calibration effects, estimation error,
and relative SNR drift. Experiments across LLaMA, GPT-2, and Qwen
show improvements over the corresponding base optimizers, including
pre-training models up to 7B parameters. \name{} also improves
fine-tuning accuracy and integrates with Adam-mini, 8-bit Adam,
and LoRA. It approaches the best manually searched module-wise
configuration on LLaMA-130M and adds less than 2\% end-to-end
training time in the 1.3B comparisons. Together, these results
establish gradient SNR as an effective basis for automated
module-wise learning-rate calibration.

%% file: Sections/appendix.tex
\clearpage
\onecolumn
\appendices
\raggedbottom

\section{Experimental Details}\label{app:experimental_details}
This appendix gives the model configurations, optimizer
hyperparameters, and implementation details.
For Adam, we use $(\beta_1,\beta_2,\epsilon)=(0.9,0.95,1\mathrm{e}{-8})$
across the reported tasks, following common LLM training
settings~\cite{touvron2023llama,wang2025sharpness}.
The pre-training learning-rate grid is
$\{1\mathrm{e}{-4},2.5\mathrm{e}{-4},5\mathrm{e}{-4},1\mathrm{e}{-3},2.5\mathrm{e}{-3},5\mathrm{e}{-3},1\mathrm{e}{-2},2.5\mathrm{e}{-2}\}$.
Tables~\ref{tab:lr_parameter} and~\ref{tab:lr_parameter_mem_efficient}
list the selected values for full-parameter and memory-efficient
methods.

Standard pre-training uses sequence length 256 and total batch size
512, or 131,072 tokens per iteration.
We use BF16 precision and gradient clipping at 1.0.
Linear warmup covers the first 10\% of training, followed by cosine
decay to 10\% of the initial learning rate.
Training uses Distributed Data Parallel (DDP) through PyTorch's
\texttt{torch.distributed} framework~\cite{paszke2017pytorch}.
The main multi-run comparisons use seeds 0--2, with seed 42 for
data shuffling. Gradient checkpointing is enabled for LLaMA-3B and 7B.
Table~\ref{tab:llama_parameter} lists model architectures and nominal
token budgets.

For \name{}, SNR calibration takes place after approximately the
first 1\% of planned optimizer updates. The estimated multipliers
are introduced over the remaining original warmup steps and reach
their targets at its 10\% endpoint.

Common-sense fine-tuning follows Liu et al.~\cite{liu2024dora}
and Zhu et al.~\cite{zhu2024apollosgdlikememoryadamwlevel}.
Table~\ref{tab:mmul_used_lr} gives the MMLU/LoRA settings.

\begin{table}[!htbp]
    \centering
    \resizebox{0.6\linewidth}{!}{
    \begin{tabular}{lcccccc}
    \toprule
    Methods & \textbf{60M} & \textbf{130M} & \textbf{350M} & \textbf{1B} & \textbf{3B} & \textbf{7B} \\
    \midrule
    Adam (8-bit) & 5.0e-3 & 1.0e-3 & 1.0e-3 & 5.0e-4 & 5.0e-4 & 5.0e-4 \\
    NAdam & 5.0e-3 & 1.0e-3 & 1.0e-3 & 5.0e-4 & -- & -- \\
    LAMB & 1.0e-2 & 1.0e-2 & 5.0e-3 & 5.0e-3 & -- & -- \\
    Muon & 5.0e-3 & 2.5e-3 & 1.0e-3 & 1.0e-3 & 1.0e-3 & 5.0e-4 \\
    Adam (8-bit)+\name & 2.5e-3 & 2.5e-3 & 1.0e-3 & 1.0e-3 & 5.0e-4 & 5.0e-4 \\
    \bottomrule
    \end{tabular}
    }
    \caption{Selected pre-training base learning rates for LLaMA models.}
    \label{tab:lr_parameter}
\end{table}

\begin{table}[!htbp]
    \centering
    \resizebox{0.6\linewidth}{!}{
    \begin{tabular}{l|ccc}
    \toprule
    \textbf{Hyperparameter} & Gemma3-1B & LLaMA3.2-3B & Qwen2.5-7B \\
    \midrule
    LR scheduler & \multicolumn{3}{c}{Cosine} \\
    LoRA rank $r$ & 4 & 8 & 8 \\
    Warmup steps & \multicolumn{3}{c}{10\%} \\
    Epochs & \multicolumn{3}{c}{3} \\
    Batch size & \multicolumn{3}{c}{16} \\
    Evaluation shots & \multicolumn{3}{c}{5} \\
    LoRA target modules & \multicolumn{3}{c}{All} \\
    Cutoff length & \multicolumn{3}{c}{2048} \\
    Learning rate & 2.0e-4 & 2.5e-5 & 1.0e-5 \\
    \bottomrule
    \end{tabular}
    }
    \caption{Reported fine-tuning/evaluation settings for MMLU.}
    \label{tab:mmul_used_lr}
\end{table}

\begin{table}[!htbp]
    \centering
    \resizebox{\linewidth}{!}{%
    \begin{tabular}{l|cc|cc|cc|cc|cc|cc}
    \toprule
    Models & \multicolumn{2}{c|}{\textbf{60M}} & \multicolumn{2}{c|}{\textbf{130M}} & \multicolumn{2}{c|}{\textbf{350M}} & \multicolumn{2}{c|}{\textbf{1B}} & \multicolumn{2}{c|}{\textbf{3B}} & \multicolumn{2}{c}{\textbf{7B}} \\
    \midrule
    Hyperparams & $lr$ & $\alpha$ & $lr$ & $\alpha$ & $lr$ & $\alpha$ & $lr$ & $\alpha$ & $lr$ & $\alpha$ & $lr$ & $\alpha$ \\
    \midrule
    Adam-mini & 5.0e-3 & -- & 1.0e-3 & -- & 5.0e-4 & -- & 2.5e-4 & -- & -- & -- & -- & -- \\
    GaLore & 1.0e-2 & 0.25 & 1.0e-2 & 0.25 & 1.0e-2 & 0.25 & 1.0e-2 & 0.25 & 5.0e-3 & 0.25 & 1.0e-2 & 0.25 \\
    APOLLO & 1.0e-2 & 1.0 & 1.0e-2 & 1.0 & 1.0e-2 & 1.0 & 1.0e-2 & 1.0 & 5.0e-3 & 1.0 & 1.0e-2 & 1.0 \\
    Adam-mini+\name & 2.5e-3 & -- & 1.0e-3 & -- & 1.0e-3 & -- & 1.0e-3 & -- & -- & -- & -- & -- \\
    \bottomrule
    \end{tabular}%
    }
    \caption{Selected learning rates $lr$ and method-specific $\alpha$ parameters for memory-efficient LLaMA pre-training.}
    \label{tab:lr_parameter_mem_efficient}
\end{table}

\begin{table}[!htbp]
    \centering
    \resizebox{.6\linewidth}{!}{
    \begin{tabular}{l|ccccccc}
    \toprule
    Model & Params & Hidden & Intermediate & Heads & Layers & Iterations & Training tokens \\
    \midrule
    \multirow{6}{*}{LLaMA}
    & 60M & 512 & 1376 & 8 & 8 & 10K & 1.3B \\
    & 130M & 768 & 2048 & 12 & 12 & 20K & 2.6B \\
    & 350M & 1024 & 2736 & 16 & 24 & 60K & 7.8B \\
    & 1B & 2048 & 5461 & 32 & 24 & 100K & 13.1B \\
    & 3B & 2560 & 6848 & 32 & 32 & 120K & 15.7B \\
    & 7B & 4096 & 11008 & 32 & 32 & 150K & 19.7B \\
    \midrule
    Qwen & 350M & 1024 & 3328 & 16 & 26 & 60K & 7.8B \\
    \midrule
    \multirow{2}{*}{GPT}
    & 125M & 768 & 3072 & 12 & 12 & 20K & 2.6B \\
    & 355M & 1024 & 4096 & 16 & 24 & 60K & 7.8B \\
    \bottomrule
    \end{tabular}
    }
    \caption{Architecture hyperparameters and nominal pre-training schedules for LLaMA, Qwen, and GPT models.}
    \label{tab:llama_parameter}
\end{table}

\section{Additional SNR Estimation Details}\label{app:snr_details}
Let $T$ be the planned number of optimizer updates and
$T_{\mathrm w}$ the original warmup length.
We perform one-shot calibration at $t_0\approx0.01T$, after all
modules have followed the same base learning-rate schedule.
In the standard pre-training setup, $T_{\mathrm w}=0.10T$.
The target multipliers are therefore introduced between approximately
1\% and 10\% of training, within the existing warmup interval.

At $\mathbf w^{t_0}$, the default calibration collects 32 mini-batch
gradients, each using 512 training examples, for 16,384 examples
in total. Parameters are held fixed during collection, and neither
the optimizer-update count nor the learning-rate schedule advances.
We average coordinate-wise SNR estimates within each tensor and
then across tensors in a module group.
The resulting $\widehat S_m$ values determine one set of target
learning-rate multipliers.

For $N$ collected mini-batch gradients at the calibration state,
we use sample means and unbiased sample variances,
\begin{equation}
\nonumber
 \widehat\mu_{m,i}=\frac1N\sum_{j=1}^N g_{m,i}^{(j)},\qquad
 \widehat\sigma_{m,i}^{\,2}=\frac1{N-1}\sum_{j=1}^N
       (g_{m,i}^{(j)}-\widehat\mu_{m,i})^2,
\end{equation}
and the tensor-wise aggregation weights $\omega_{m,i}$ give
\begin{equation}
\nonumber
 \widehat S_m=\sum_i\omega_{m,i}
       \frac{\widehat\mu_{m,i}^{\,2}}
            {\widehat\sigma_{m,i}^{\,2}+\epsilon_{\mathrm{SNR}}}.
\end{equation}
Appendix~\ref{app:mols_sampling} analyzes this stabilized estimator.
Each observation is the gradient of the complete mini-batch loss,
including all predicted token positions.

\begin{algorithm}[!htbp]
\caption{SNR-based module-wise learning-rate scaling}
\label{alg:snr_lr_simple}
\begin{algorithmic}
\STATE \textbf{Input:} Module groups $\mathcal M$; reference module
$m_{\mathrm{base}}$; planned optimizer updates $T$; original global
schedule $\{\eta_t\}$ with warmup length $T_{\mathrm w}$;
calibration step $t_0<T_{\mathrm w}$ (default $t_0\approx0.01T$);
calibration budget $N$; SNR stabilizer $\epsilon_{\mathrm{SNR}}$.
\STATE Run the base optimizer with the same learning-rate schedule
for all groups until $\mathbf w^{t_0}$.
\STATE Hold the parameters and schedule step fixed. Collect $N$
mini-batch gradients and estimate $\{\widehat S_m\}_{m\in\mathcal M}$
using the stated aggregation and stabilizer $\epsilon_{\mathrm{SNR}}$.
\STATE $\widehat S_{\mathrm{base}}\leftarrow\widehat S_{m_{\mathrm{base}}}$.
\FOR{$m\in\mathcal M$}
    \STATE $\widehat\alpha_m\leftarrow
        \sqrt{\widehat S_{\mathrm{base}}/\widehat S_m}$.
\ENDFOR
\STATE Resume optimizer updates. Over the original interval from
$t_0$ to $T_{\mathrm w}$, gradually transition each applied multiplier
from $1$ to its target $\widehat\alpha_m$.
\STATE At $T_{\mathrm w}$, finish warmup on the original schedule.
For subsequent updates, keep the multipliers fixed and use
$\eta_{m,t}=\widehat\alpha_m\eta_t$ with the original decay schedule.
\end{algorithmic}
\end{algorithm}

\paragraph{Normalization and warmup transition}
The LLaMA experiments use VO as the reference, with
$\widehat\alpha_{\mathrm{VO}}=1$.
Each target multiplier is computed once at $t_0$.
The applied multiplier starts at $1$ and reaches
$\widehat\alpha_m$ at the original warmup endpoint
$T_{\mathrm w}$, then remains fixed.
The transition avoids an abrupt change in learning rate at
calibration and requires no additional warmup stage.
The original warmup length, subsequent decay, and optimizer-update
budget are preserved.
Table~\ref{tab:alpha_value} reports the estimates for 60M--7B models
using the notation $\alpha_m$.

\section{Proofs}
\label{app:mols_proofs}


In a group of $K_m$ tensors, a coordinate in tensor $p$ of size
$d_p$ has weight $\omega_{m,i}=1/(K_m d_p)$.
Let $d_m$ be the total number of group coordinates.
Write $\mathbf d_m^t=(d_{m,i}^t)_{i=1}^{d_m}$ with
$d_{m,i}^t=\mu_{m,i}^t/\sqrt{(\mu_{m,i}^t)^2+(\sigma_{m,i}^t)^2}$,
and set $\mathbf W_m=\operatorname{diag}(\omega_{m,1},\ldots,\omega_{m,d_m})$.
The signal quantity in Eq.~(\ref{eq:adam_step_snr}) satisfies
\begin{equation}
 (R_m^t)^2 =\|\mathbf W_m^{1/2}\mathbf d_m^t\|_2^2 =\sum_i\omega_{m,i}\frac{s_{m,i}^t}{1+s_{m,i}^t}.
 \nonumber
\end{equation}
We abbreviate $\mathcal K_t(\mathbf1)$,
$\mathcal K_t(\mathbf a^t)$, and
$\mathcal K_t(\widehat{\mathbf a})$ by
$\mathcal K_{\mathrm{plain}}^t$,
$\mathcal K_{\mathrm{cal}}^t$, and
$\mathcal K_{\mathrm{fixed}}^t$, respectively.
The reference group $b$ may lie inside or outside $\mathcal G$.

The signal bounds use the population-moment model, as in
moment-based analyses of Adam~\cite{balles2018dissecting}.
Theorem~\ref{thm:mols_one_shot} concerns score ratios, while
Proposition~\ref{prop:mols_structure} uses raw-gradient moments.
The numerical SNR stabilizer is retained in the finite-sample analysis.
Assumption~\ref{ass:mols_moments} applies at each state; coordinate
means may be zero, and moments may change with $t$.

\subsection{Proof of Lemma~\ref{lem:mols_signal}}

\begin{molslemma}[Aggregation with heterogeneous coordinate SNRs]
\label{lem:mols_signal}
Under Assumption~\ref{ass:mols_moments},
\begin{equation}
 \frac{S_m^t}{1+\tau_m^t} \leq (R_m^t)^2 \leq \frac{S_m^t}{1+S_m^t}.
 \label{eq:mols_signal_sandwich}
\end{equation}
Moreover,
\begin{equation}
 \frac{S_m^t}{1+S_m^t}-(R_m^t)^2 =\frac{1}{(1+S_m^t)^2}
 \sum_i\frac{\omega_{m,i}(s_{m,i}^t-S_m^t)^2}{1+s_{m,i}^t}.
 \label{eq:mols_aggregation_gap}
\end{equation}
\end{molslemma}

\begin{proof}
Fix $m,t$ and suppress these indices. Write $s_i=s_{m,i}^t$,
$\omega_i=\omega_{m,i}$, $S=S_m^t$, $R=R_m^t$, and
$\tau=\tau_m^t$. Since $S=\sum_i\omega_i s_i>0$ and
$R^2=\sum_i\omega_i s_i/(1+s_i)$,
\begin{align}
 \frac{S}{1+S}-R^2 &=\sum_i\omega_i\left(\frac{1}{1+s_i}-\frac{1}{1+S}\right)
 \overset{\mathrm{(a)}}{=} \sum_i\omega_i\left( \frac{1}{1+s_i}-\frac{1}{1+S}
 +\frac{s_i-S}{(1+S)^2}\right)
 \nonumber\\
 &=\frac{1}{(1+S)^2} \sum_i\frac{\omega_i(s_i-S)^2}{1+s_i} \geq0,
 \nonumber
\end{align}
where (a) uses $\sum_i\omega_i(s_i-S)=0$.
This gives Eq.~(\ref{eq:mols_aggregation_gap}) and the upper bound
in Eq.~(\ref{eq:mols_signal_sandwich}). For the lower bound,
\begin{align}
 S^2 &=\left(\sum_i \sqrt{\frac{\omega_i s_i}{1+s_i}} \sqrt{\omega_i s_i(1+s_i)}\right)^2
 \nonumber\\
 &\overset{\mathrm{(b)}}{\leq} \left(\sum_i\frac{\omega_i s_i}{1+s_i}\right) \left(\sum_i\omega_i
 s_i(1+s_i)\right) =R^2\left(S+\sum_i\omega_i s_i^2\right) =R^2S(1+\tau),
 \nonumber\\
 \tau &=\frac{\sum_i\omega_i s_i^2}{S} =S+\frac{\sum_i\omega_i(s_i-S)^2}{S} \geq S.
 \nonumber
\end{align}
Here (b) is Cauchy--Schwarz. Dividing the first chain by
$S(1+\tau)>0$ proves the lower bound. Both bounds coincide
when $s_i=S$ for every $i$. The proof is completed.
\end{proof}

\subsection{Proof of Theorem~\ref{thm:mols_calibration}}

\begin{proof}
Fix $t$. Multiplying Eq.~(\ref{eq:mols_signal_sandwich}) by
$(\alpha_m^t)^2=S_b^t/S_m^t$ gives
$S_b^t/(1+\tau_m^t)\leq(\alpha_m^tR_m^t)^2
\leq S_b^t/(1+S_m^t)$.
Taking square roots yields Eq.~(\ref{eq:mols_calibrated_sandwich}).
Consequently,
\begin{equation}
 \mathcal K_{\mathrm{cal}}^t =\frac{\max_{m\in\mathcal G}\alpha_m^tR_m^t} {\min_{m\in\mathcal
 G}\alpha_m^tR_m^t} \leq \frac{\sqrt{S_b^t/(1+S_-^t)}} {\sqrt{S_b^t/(1+\tau_+^t)}}
 =\sqrt{\frac{1+\tau_+^t}{1+S_-^t}} =H^t,
 \nonumber
\end{equation}
where $H^t\geq1$ follows from $\tau_+^t\geq S_+^t\geq S_-^t$.
Choose $m_+,m_-\in\mathcal G$ with scores $S_+^t,S_-^t$.
Applying Lemma~\ref{lem:mols_signal} to these two groups gives
\begin{align}
 \mathcal K_{\mathrm{plain}}^t &=\frac{\max_{m\in\mathcal G}R_m^t} {\min_{m\in\mathcal G}R_m^t}
 \geq\frac{R_{m_+}^t}{R_{m_-}^t} \geq\sqrt{ \frac{S_+^t}{1+\tau_{m_+}^t} \frac{1+S_-^t}{S_-^t}}
 \geq\sqrt{ \frac{S_+^t}{S_-^t} \frac{1+S_-^t}{1+\tau_+^t}} =\frac{\sqrt{S_+^t/S_-^t}}{H^t},
 \nonumber\\
 \frac{\mathcal K_{\mathrm{cal}}^t}{\mathcal K_{\mathrm{plain}}^t}
 &\leq\frac{H^t}{\sqrt{S_+^t/S_-^t}/H^t} =\frac{(H^t)^2}{\sqrt{S_+^t/S_-^t}}.
 \nonumber
\end{align}
The last expression is below one when $S_+^t/S_-^t>(H^t)^4$.
The reference factor $\sqrt{S_b^t}$ cancels from the ratios, whether
or not $b\in\mathcal G$.
For homogeneous coordinates in a group,
$R_m^t=\sqrt{S_m^t/(1+S_m^t)}$ and
$\alpha_m^tR_m^t=\sqrt{S_b^t/(1+S_m^t)}$, which gives the
remaining dependence on module SNR. The proof is completed.
\end{proof}

\subsection{Proof of Theorem~\ref{thm:mols_one_shot}}

\begin{proof}
Write $\operatorname{osc}(\mathbf z)=\max_m z_m-\min_m z_m$.
For positive $\mathbf x,\mathbf y$, setting
$z_m=\log x_m-\log y_m$ gives
$d_{\mathrm{rel}}(\mathbf x,\mathbf y)
=\max_{m,n}|z_m-z_n|=\operatorname{osc}(\mathbf z)$.
Also, $\operatorname{osc}(\mathbf z+c\mathbf1)
=\operatorname{osc}(-\mathbf z)=\operatorname{osc}(\mathbf z)$.
Define $e_m=\log(\widehat S_m/S_m^{t_0})$ and
$r_m^t=\log(S_m^t/S_m^{t_0})$, with
$E=\tfrac12\operatorname{osc}(\mathbf e)$ and
$D^t=\tfrac12\operatorname{osc}(\mathbf r^t)$.
The multiplier ratios satisfy
\begin{align}
 \log\frac{\widehat\alpha_m/\widehat\alpha_n} {\alpha_m^t/\alpha_n^t} &=\frac12\log\frac{\widehat
 S_n/\widehat S_m}{S_n^t/S_m^t} =\frac12\left( \log\frac{\widehat S_n}{S_n^t} -\log\frac{\widehat
 S_m}{S_m^t}\right) =\frac12\bigl[(e_n-r_n^t)-(e_m-r_m^t)\bigr],
 \nonumber\\
 \Delta^t &=d_{\mathrm{rel}}(\widehat{\mathbf a},\mathbf a^t)
 =\frac12d_{\mathrm{rel}}(\widehat{\mathbf s},\mathbf s^t) =\frac12\operatorname{osc}(\mathbf
 e-\mathbf r^t) \overset{\mathrm{(a)}}{\leq} \frac12\operatorname{osc}(\mathbf e)
 +\frac12\operatorname{osc}(\mathbf r^t) =E+D^t.
 \nonumber
\end{align}
In (a), $\operatorname{osc}(\mathbf z+\mathbf u)
\leq\operatorname{osc}(\mathbf z)+\operatorname{osc}(\mathbf u)$.
Since $E=\tfrac12d_{\mathrm{rel}}(\widehat{\mathbf s},\mathbf s^{t_0})$
and $D^t=\tfrac12d_{\mathrm{rel}}(\mathbf s^{t_0},\mathbf s^t)$,
this proves Eq.~(\ref{eq:mols_half_sensitivity}).

Every ordered multiplier ratio lies between
$\exp(-\Delta^t)$ and $\exp(\Delta^t)$ times its population
counterpart. Thus
\begin{align}
 \mathcal K_{\mathrm{fixed}}^t &=\max_{m,n\in\mathcal G}
 \frac{\widehat\alpha_mR_m^t}{\widehat\alpha_nR_n^t} =\max_{m,n\in\mathcal G}\left(
 \frac{\widehat\alpha_m/\widehat\alpha_n}{\alpha_m^t/\alpha_n^t}
 \frac{\alpha_m^tR_m^t}{\alpha_n^tR_n^t}\right)
 \nonumber\\
 &\leq\exp(\Delta^t)\max_{m,n\in\mathcal G} \frac{\alpha_m^tR_m^t}{\alpha_n^tR_n^t}
 =\exp(\Delta^t)\mathcal K_{\mathrm{cal}}^t \overset{\mathrm{(b)}}{\leq}\exp(\Delta^t)H^t
 \leq\exp(E+D^t)H^t,
 \nonumber
\end{align}
where (b) follows from Theorem~\ref{thm:mols_calibration}.
All denominators are positive by Assumption~\ref{ass:mols_moments}.
If $\widehat S_m=c_0S_m^{t_0}$ and $S_m^t=c_tS_m^{t_0}$ for
all $m$, then $\mathbf e=(\log c_0)\mathbf1$ and
$\mathbf r^t=(\log c_t)\mathbf1$. Hence $E=D^t=0$ and
$\widehat\alpha_m=\alpha_m^t=\sqrt{S_b^{t_0}/S_m^{t_0}}$.
The proof is completed.
\end{proof}

\subsection{Proof of Corollary~\ref{cor:mols_finite_sample}}
\label{app:mols_sampling}

\begin{molsassumption}[Local calibration sampling]
\label{ass:mols_sampling}
For the finite-sample result only, calibration uses $N\geq2$
independent, identically distributed mini-batch gradients evaluated
at a fixed parameter state $\mathbf w^{t_0}$. Their coordinate-wise
centered fourth moments are finite. Coordinates within each gradient
vector may be dependent.
\end{molsassumption}

Finite fourth moments control the sample-variance error, as in Adam
analyses~\cite{barakat2021convergence}. This condition holds for
finite-support calibration gradients with finite values.

\begin{molscorollary}[Finite-sample consistency of calibration]
\label{cor:mols_finite_sample}
Under Assumptions~\ref{ass:mols_moments} and~\ref{ass:mols_sampling},
fix $\lambda=\epsilon_{\mathrm{SNR}}>0$ and let
\begin{equation}
 \widehat S_m =\sum_i\omega_{m,i} \frac{\widehat\mu_{m,i}^{\,2}}
 {\widehat\sigma_{m,i}^{\,2}+\lambda}, \qquad S_{m,\lambda} =\sum_i\omega_{m,i}
 \frac{(\mu_{m,i}^{t_0})^2} {(\sigma_{m,i}^{t_0})^2+\lambda},
 \nonumber
\end{equation}
where $\widehat\mu_{m,i}$ and $\widehat\sigma_{m,i}^{\,2}$ are the
sample mean and unbiased sample variance of the $N$ gradients.
There is a finite constant $C$, given in
Eq.~(\ref{eq:mols_sampling_constant}) and independent of $N,\delta$,
such that, for $0<\delta<1$ and
$r_N=C/(\delta\sqrt N)\leq1/2$, with probability at least
$1-\delta$ all $\widehat S_m$ are positive and
\begin{equation}
 \max_m\frac{|\widehat S_m-S_{m,\lambda}|}{S_{m,\lambda}} \leq r_N, \qquad
 d_{\mathrm{rel}}(\widehat{\mathbf a},\mathbf a_\lambda) \leq\frac12\log\frac{1+r_N}{1-r_N}
 \leq2r_N,
 \label{eq:mols_finite_sample_bound}
\end{equation}
where $(\mathbf a_\lambda)_m=\sqrt{S_{b,\lambda}/S_{m,\lambda}}$.
\end{molscorollary}

The rate is $N^{-1/2}$ at fixed confidence and stabilizer.
The constant depends on the local moments and population scores.
Table~\ref{tab:snr_samples_ppl} reports the practical calibration budgets.

\begin{proof}
All population quantities are evaluated at $t_0$.
Fix a coordinate and suppress its indices. For calibration gradients
$G^1,\ldots,G^N$, write $X^j=G^j-\mu$, so that
$\mathbb E X^j=0$, $\mathbb E(X^j)^2=\sigma^2$, and
$\mathbb E(X^j)^4=\kappa<\infty$.
Set $U_N=\sum_jX^j$ and $V_N=\sum_j(X^j)^2$.
Expanding the moments gives
\begin{align}
 \mathbb E U_N^2 &=\sum_j\mathbb E(X^j)^2 =N\sigma^2=\mathbb E V_N,
 \nonumber\\
 \mathbb E V_N^2 &=\sum_j\mathbb E(X^j)^4 +2\sum_{j<k}\mathbb E(X^j)^2\mathbb E(X^k)^2
 =N\kappa+N(N-1)\sigma^4,
 \nonumber\\
 \mathbb E U_N^4 &=\sum_j\mathbb E(X^j)^4 +6\sum_{j<k}\mathbb E(X^j)^2\mathbb E(X^k)^2
 =N\kappa+3N(N-1)\sigma^4,
 \nonumber\\
 \mathbb E(U_N^2V_N) &=\mathbb E V_N^2 +2\sum_{j<k}\sum_\ell\mathbb E[X^jX^k(X^\ell)^2]
 \overset{\mathrm{(a)}}{=}N\kappa+N(N-1)\sigma^4.
 \nonumber
\end{align}
Independence removes the terms with an unpaired centered factor.
In particular, each term in the triple sum in (a) contains such a
factor, including when $\ell=j$ or $\ell=k$.
The sample identities
$\widehat\mu-\mu=U_N/N$ and
$\widehat\sigma^{\,2}=(V_N-U_N^2/N)/(N-1)\geq0$ then yield
\begin{align}
 \mathbb E\widehat\sigma^{\,2} &=\frac{\mathbb E V_N-\mathbb E U_N^2/N}{N-1} =\sigma^2,
 \nonumber\\
 \mathbb E(\widehat\sigma^{\,2})^2 &=\frac{\mathbb E V_N^2 -2\mathbb E(V_NU_N^2)/N+\mathbb E
 U_N^4/N^2}{(N-1)^2} =\frac\kappa N +\frac{N^2-2N+3}{N(N-1)}\sigma^4,
 \nonumber\\
 \operatorname{Var}(\widehat\sigma^{\,2}) &=\frac1N\left(\kappa-\frac{N-3}{N-1}\sigma^4\right)
 \leq\frac{\kappa+\sigma^4}{N},
 \nonumber\\
 \mathbb E|\widehat\mu^{\,2}-\mu^2| &=\mathbb E|2\mu(\widehat\mu-\mu)+(\widehat\mu-\mu)^2|
 \overset{\mathrm{(b)}}{\leq} 2|\mu|\sqrt{\mathbb E(\widehat\mu-\mu)^2} +\mathbb
 E(\widehat\mu-\mu)^2 =\frac{2|\mu|\sigma}{\sqrt N}+\frac{\sigma^2}{N}.
 \nonumber
\end{align}
The variance bound holds for $N\geq2$, and (b) follows from the
triangle inequality and Cauchy--Schwarz.
For $s_\lambda=\mu^2/(\sigma^2+\lambda)$ and
$\widehat s_\lambda=\widehat\mu^{\,2}/(\widehat\sigma^{\,2}+\lambda)$,
\begin{align}
 \widehat s_\lambda-s_\lambda &=\frac{\widehat\mu^{\,2}-\mu^2
 -s_\lambda(\widehat\sigma^{\,2}-\sigma^2)} {\widehat\sigma^{\,2}+\lambda},
 \nonumber\\
 \mathbb E|\widehat s_\lambda-s_\lambda| &\overset{\mathrm{(c)}}{\leq}\frac1\lambda \left(\mathbb
 E|\widehat\mu^{\,2}-\mu^2| +s_\lambda\mathbb E|\widehat\sigma^{\,2}-\sigma^2|\right)
 \nonumber\\
 &\leq\frac1\lambda\left( \frac{2|\mu|\sigma}{\sqrt N}+\frac{\sigma^2}{N}
 +s_\lambda\sqrt{\frac{\kappa+\sigma^4}{N}}\right) \leq\frac{2|\mu|\sigma+\sigma^2
 +s_\lambda\sqrt{\kappa+\sigma^4}} {\lambda\sqrt N},
 \nonumber
\end{align}
where (c) uses $\widehat\sigma^{\,2}+\lambda\geq\lambda$.
The next line uses the unbiasedness and variance bound of
$\widehat\sigma^{\,2}$.

Restore the indices and abbreviate
$\mu_{m,i}=\mu_{m,i}^{t_0}$,
$\sigma_{m,i}=\sigma_{m,i}^{t_0}$, and
$\kappa_{m,i}=\mathbb E[(g_{m,i}^{t_0}-\mu_{m,i}^{t_0})^4]$.
With $s_{m,i,\lambda}=\mu_{m,i}^2/(\sigma_{m,i}^2+\lambda)$, define
\begin{equation}
 C_m =\frac1\lambda\sum_i\omega_{m,i} \left(2|\mu_{m,i}|\sigma_{m,i}+\sigma_{m,i}^2
 +s_{m,i,\lambda}\sqrt{\kappa_{m,i}+\sigma_{m,i}^4}\right), \qquad C=\sum_{m\in\mathcal
 M}\frac{C_m}{S_{m,\lambda}}.
 \label{eq:mols_sampling_constant}
\end{equation}
These constants are finite, and $S_{m,\lambda}>0$ by
Assumption~\ref{ass:mols_moments}.
Summing the coordinate bounds and applying Markov's inequality gives
\begin{align}
 \mathbb E|\widehat S_m-S_{m,\lambda}| &\leq\sum_i\omega_{m,i} \mathbb E|\widehat
 s_{m,i,\lambda}-s_{m,i,\lambda}| \leq\frac{C_m}{\sqrt N},
 \nonumber\\
 \mathbb P\!\left( \max_m\frac{|\widehat S_m-S_{m,\lambda}|}{S_{m,\lambda}}>r_N\right)
 &\leq\frac1{r_N}\mathbb E\left[ \sum_m\frac{|\widehat S_m-S_{m,\lambda}|}{S_{m,\lambda}}\right]
 \leq\frac{C}{r_N\sqrt N}=\delta.
 \nonumber
\end{align}
On the complementary event,
$1-r_N\leq\widehat S_m/S_{m,\lambda}\leq1+r_N$ for every $m$.
Since $r_N\leq1/2$, all empirical scores are positive.
Put $z_m=\log(\widehat S_m/S_{m,\lambda})$.
Equation~(\ref{eq:mols_half_sensitivity}) then gives
\begin{align}
 d_{\mathrm{rel}}(\widehat{\mathbf a},\mathbf a_\lambda) &=\frac12\bigl(\max_m z_m-\min_m z_m\bigr)
 \leq\frac12\bigl[\log(1+r_N)-\log(1-r_N)\bigr]
 \nonumber\\
 &=\int_0^{r_N}\frac{\mathrm du}{1-u^2} \leq\frac{r_N}{1-r_N^2} \leq2r_N.
 \nonumber
\end{align}
This proves Eq.~(\ref{eq:mols_finite_sample_bound}).

To separate stabilization bias, set
$z_{m,\lambda}^{\mathrm{bias}}=\log(S_{m,\lambda}/S_m^{t_0})$.
Direct subtraction gives
\begin{equation}
 S_m^{t_0}-S_{m,\lambda} =\sum_i\omega_{m,i}\mu_{m,i}^2 \left(\frac1{\sigma_{m,i}^2}
 -\frac1{\sigma_{m,i}^2+\lambda}\right) =\sum_i\omega_{m,i}s_{m,i}^{t_0}
 \frac\lambda{\sigma_{m,i}^2+\lambda} \geq0.
 \nonumber
\end{equation}
Using $e_m=z_m+z_{m,\lambda}^{\mathrm{bias}}$ and
Theorem~\ref{thm:mols_one_shot}, on the same event we obtain
\begin{align}
 E &=\tfrac12\operatorname{osc} (\mathbf z+\mathbf z_\lambda^{\mathrm{bias}})
 \leq\frac12\log\frac{1+r_N}{1-r_N} +\frac12\operatorname{osc}(\mathbf z_\lambda^{\mathrm{bias}}),
 \nonumber\\
 \mathcal K_{\mathrm{fixed}}^t &\leq H^t\exp(E+D^t) \leq H^t\sqrt{\frac{1+r_N}{1-r_N}}
 \exp\!\left(D^t+ \frac12\operatorname{osc}(\mathbf z_\lambda^{\mathrm{bias}})\right).
 \nonumber
\end{align}
The last bound holds at every later state satisfying
Assumption~\ref{ass:mols_moments}, with the same fixed $\lambda>0$.
The proof is completed.
\end{proof}

\subsection{Proof of Proposition~\ref{prop:mols_structure}}
\label{app:mols_structural_proof}

\begin{proof}
Fix $\mathbf w^t,\mathbf x$ as in the proposition and suppress
these arguments. Expectations and covariances are conditional on
$\mathbf x$. The embedding $\mathbf E$ and head $\mathbf U$ are
untied. Use the tensor weights defined at the start of this appendix,
and assign score zero to identically zero gradient coordinates.
The internal groups are QK, VO, and MLP across all layers.

Let $\mathbf e_j$ be the $j$-th standard basis vector,
$\boldsymbol\delta=\mathbf p-\mathbf e_y$, and
$\mathbf b_{m,i}=\partial\mathbf z/\partial w_{m,i}$.
The chain rule and the distribution $y\sim\mathbf q$ give
\begin{align}
 g_{m,i} &=\frac{\partial\ell}{\partial w_{m,i}} =\mathbf b_{m,i}^\top\boldsymbol\delta,
 \nonumber\\
 \mathbb E[\boldsymbol\delta\mid\mathbf x] &=\mathbf p-\sum_jq_j\mathbf e_j =\mathbf p-\mathbf
 q=\mathbf r,
 \nonumber\\
 \operatorname{Cov}(\boldsymbol\delta\mid\mathbf x) &=\sum_jq_j(\mathbf e_j-\mathbf q)(\mathbf
 e_j-\mathbf q)^\top =\operatorname{diag}(\mathbf q)-\mathbf q\mathbf q^\top =\mathbf C,
 \nonumber\\
 \mathbf v^\top\mathbf C\mathbf v &=\sum_jq_jv_j^2-\left(\sum_jq_jv_j\right)^2
 =\frac12\sum_{j,k}q_jq_k(v_j-v_k)^2 \geq0.
 \nonumber
\end{align}
Because every $q_j>0$, the final quadratic form vanishes exactly
when $\mathbf v$ is constant. Hence
$\ker(\mathbf C)=\operatorname{span}\{\mathbf1_V\}$ and
$\operatorname{range}(\mathbf C)=\mathbf1_V^\perp$.
In particular, $\mathbf r\in\operatorname{range}(\mathbf C)$.
With $\mathbf D=\operatorname{diag}(\mathbf q)$,
\begin{align}
 \mathbf C\mathbf D^{-1}\mathbf r &=\mathbf r-\mathbf q\mathbf1_V^\top\mathbf r =\mathbf r, \qquad
 \mathbf C(\mathbf C^\dagger\mathbf r-\mathbf D^{-1}\mathbf r) =\mathbf r-\mathbf r=\mathbf0,
 \nonumber\\
 \Gamma &=\mathbf r^\top\mathbf C^\dagger\mathbf r \overset{\mathrm{(a)}}{=}\mathbf r^\top\mathbf
 D^{-1}\mathbf r =\sum_j\frac{r_j^2}{q_j}.
 \nonumber
\end{align}
Equality (a) follows because
$\mathbf C^\dagger\mathbf r-\mathbf D^{-1}\mathbf r$ lies in
$\ker(\mathbf C)$ and is orthogonal to $\mathbf r$.

Define $\mathbf u=(\mathbf C^\dagger)^{1/2}\mathbf r/\sqrt\Gamma$.
For $\mathbf b_{m,i}^\top\mathbf C\mathbf b_{m,i}>0$, put
$\mathbf v_{m,i}=\mathbf C^{1/2}\mathbf b_{m,i}/
\sqrt{\mathbf b_{m,i}^\top\mathbf C\mathbf b_{m,i}}$;
otherwise put $\mathbf v_{m,i}=\mathbf0$.
Let $\chi_{m,i}=(\mathbf v_{m,i}^\top\mathbf u)^2$ and
$\kappa_m=\sum_i\omega_{m,i}\chi_{m,i}$.
The nonzero vectors in these definitions are unit vectors, so
$0\leq\chi_{m,i}\leq1$. For a positive-variance coordinate,
\begin{align}
 s_{m,i}(\mathbf x) &=\frac{(\mathbb E[g_{m,i}\mid\mathbf x])^2}
 {\operatorname{Var}(g_{m,i}\mid\mathbf x)} =\frac{(\mathbf b_{m,i}^\top\mathbf r)^2} {\mathbf
 b_{m,i}^\top\mathbf C\mathbf b_{m,i}}
 \nonumber\\
 &\overset{\mathrm{(b)}}{=} \frac{\bigl[(\mathbf C^{1/2}\mathbf b_{m,i})^\top (\mathbf
 C^\dagger)^{1/2}\mathbf r\bigr]^2} {\mathbf b_{m,i}^\top\mathbf C\mathbf b_{m,i}} =\Gamma(\mathbf
 v_{m,i}^\top\mathbf u)^2 =\Gamma\chi_{m,i}.
 \nonumber
\end{align}
Here (b) uses
$\mathbf C^{1/2}(\mathbf C^\dagger)^{1/2}
=\mathbf I_V-\mathbf1_V\mathbf1_V^\top/V$ and
$\mathbf1_V^\top\mathbf r=0$.
If the variance is zero, then $\mathbf b_{m,i}=c\mathbf1_V$,
so $g_{m,i}=c\mathbf1_V^\top(\mathbf p-\mathbf e_y)=0$ for
every label. Thus $s_{m,i}=\Gamma\chi_{m,i}$ also holds in this
case under the zero-score convention. Averaging gives, for
$m\in\{\mathrm{QK},\mathrm{VO},\mathrm{MLP}\}$,
\begin{equation}
 S_m(\mathbf x) =\sum_i\omega_{m,i}s_{m,i}(\mathbf x) =\Gamma\sum_i\omega_{m,i}\chi_{m,i}
 =\Gamma\kappa_m.
 \nonumber
\end{equation}

Let $\mathcal A(\mathbf x)$ contain the $K$ distinct input indices.
The untied lookup gives $\partial\mathbf z/\partial E_{rk}=\mathbf0$
for $r\notin\mathcal A(\mathbf x)$; repeated occurrences contribute
to the same row derivative.
With $\kappa_{E,\mathrm{act}}=(Kd)^{-1}
\sum_{r\in\mathcal A(\mathbf x)}\sum_{k=1}^d\chi_{E,rk}$,
\begin{equation}
 S_{\mathrm{Emb}}(\mathbf x) =\frac1{Vd}\sum_{r=1}^V\sum_{k=1}^d s_{E,rk}(\mathbf x)
 =\frac\Gamma{Vd} \sum_{r\in\mathcal A(\mathbf x)}\sum_{k=1}^d\chi_{E,rk} =\Gamma\frac K
 V\kappa_{E,\mathrm{act}}.
 \nonumber
\end{equation}
For $\kappa_{E,\mathrm{act}}>0$, it follows that
$S_m(\mathbf x)/S_{\mathrm{Emb}}(\mathbf x)
=(V/K)\kappa_m/\kappa_{E,\mathrm{act}}$.

For the output head, $\partial\mathbf z/\partial U_{jk}=h_k\mathbf e_j$.
The coordinate gradient is
$g_{\mathrm{Head},jk}=h_k(p_j-\mathbf1_{\{y=j\}})$, and hence
\begin{align}
 \mathbb E[g_{\mathrm{Head},jk}\mid\mathbf x] &=h_k(p_j-q_j)=h_kr_j,
 \nonumber\\
 \operatorname{Var}(g_{\mathrm{Head},jk}\mid\mathbf x)
 &=h_k^2\operatorname{Var}(\mathbf1_{\{y=j\}}\mid\mathbf x) =h_k^2q_j(1-q_j),
 \nonumber\\
 s_{\mathrm{Head},jk}(\mathbf x) &=\frac{h_k^2r_j^2}{h_k^2q_j(1-q_j)} =\frac{r_j^2}{q_j(1-q_j)},
 \nonumber\\
 S_{\mathrm{Head}}(\mathbf x) &=\frac1{Vd}\sum_{j=1}^V\sum_{k=1}^d \frac{r_j^2}{q_j(1-q_j)}
 =\frac1V\sum_{j=1}^V\frac{r_j^2}{q_j(1-q_j)}.
 \nonumber
\end{align}
The nonzero hidden-coordinate condition permits cancellation of
$h_k^2$. This proves Eq.~(\ref{eq:mols_structural_scores}). Finally,
\begin{equation}
 \frac\Gamma V =\frac1V\sum_j\frac{r_j^2}{q_j} \leq S_{\mathrm{Head}}(\mathbf x)
 \leq\frac1{V(1-q_{\max})}\sum_j\frac{r_j^2}{q_j} =\frac\Gamma{V(1-q_{\max})}.
 \nonumber
\end{equation}
Using $S_m=\Gamma\kappa_m$ in these two bounds gives
$V(1-q_{\max})\kappa_m
\leq S_m(\mathbf x)/S_{\mathrm{Head}}(\mathbf x)
\leq V\kappa_m$. The proof is completed.
\end{proof}

\paragraph{Consequences and hidden-coordinate support}
If $\kappa_m\geq c_0\kappa_{E,\mathrm{act}}>0$, then
$S_m/S_{\mathrm{Emb}}\geq c_0V/K$.
For uniform conditional labels,
$S_{\mathrm{Head}}=\Gamma/(V-1)$ and
$S_m/S_{\mathrm{Head}}=(V-1)\kappa_m$.
If only $d_a>0$ hidden coordinates are nonzero, the other head
coordinates have zero gradients. Averaging over all $d$ coordinates
then gives
\begin{equation}
 S_{\mathrm{Head}}(\mathbf x) =\frac{d_a}{dV}\sum_j\frac{r_j^2}{q_j(1-q_j)}, \qquad \frac
 d{d_a}V(1-q_{\max})\kappa_m \leq\frac{S_m(\mathbf x)}{S_{\mathrm{Head}}(\mathbf x)} \leq\frac
 d{d_a}V\kappa_m.
 \nonumber
\end{equation}

\paragraph{Separate attention groups}
For positive conditional scores,
$S_{\mathrm{QK}}(\mathbf x)/S_{\mathrm{VO}}(\mathbf x)
=\kappa_{\mathrm{QK}}/\kappa_{\mathrm{VO}}$.
If QK and VO contain $n_{\mathrm{QK}}$ and $n_{\mathrm{VO}}$
tensors, their merged score is
\begin{equation}
 S_{\mathrm{Attn}} =\frac{n_{\mathrm{QK}}S_{\mathrm{QK}} +n_{\mathrm{VO}}S_{\mathrm{VO}}}
 {n_{\mathrm{QK}}+n_{\mathrm{VO}}}.
 \nonumber
\end{equation}
This averaging identity also holds for full-data scores.
Separate groups retain the learning-rate ratio
$\alpha_{\mathrm{QK}}/\alpha_{\mathrm{VO}}
=\sqrt{S_{\mathrm{VO}}/S_{\mathrm{QK}}}$.

\paragraph{Attention and MLP derivatives}
The logit derivatives $\mathbf b_{m,i}$ include all decoder
operations, including normalization and residual connections.
For one causal attention head, write
$\mathbf Q=\mathbf X\mathbf W_Q$,
$\mathbf K=\mathbf X\mathbf W_K$, and
$\mathbf V_a=\mathbf X\mathbf W_V$.
Let $\mathbf L=\mathbf Q\mathbf K^\top/\sqrt{d_h}+\mathbf M$,
$\mathbf A=\operatorname{softmax}_{\mathrm{row}}(\mathbf L)$,
and $\mathbf O=\mathbf A\mathbf V_a$.
The mask $\mathbf M$ is fixed. On the allowed entries,
\begin{align}
 \mathrm d\mathbf Q &=(\mathrm d\mathbf X)\mathbf W_Q+\mathbf X\mathrm d\mathbf W_Q, \qquad \mathrm
 d\mathbf K =(\mathrm d\mathbf X)\mathbf W_K+\mathbf X\mathrm d\mathbf W_K,
 \nonumber\\
 \mathrm d\mathbf V_a &=(\mathrm d\mathbf X)\mathbf W_V+\mathbf X\mathrm d\mathbf W_V, \qquad
 \mathrm d\mathbf L =\frac{(\mathrm d\mathbf Q)\mathbf K^\top +\mathbf Q(\mathrm d\mathbf
 K)^\top}{\sqrt{d_h}},
 \nonumber\\
 \mathrm d\mathbf a_\ell &=[\operatorname{diag}(\mathbf a_\ell) -\mathbf a_\ell\mathbf a_\ell^\top]
 \mathrm d\boldsymbol\ell_\ell, \qquad \mathrm d\mathbf O =(\mathrm d\mathbf A)\mathbf V_a+\mathbf
 A\mathrm d\mathbf V_a.
 \nonumber
\end{align}
Here $\mathbf a_\ell$ and $\boldsymbol\ell_\ell$ are column forms of
the corresponding rows. Masked probabilities have zero derivatives.
Multi-head attention concatenates head outputs before projection.
RoPE applies the same fixed position-dependent rotations to
$\mathrm d\mathbf Q,\mathrm d\mathbf K$ as to $\mathbf Q,\mathbf K$.

For SwiGLU, let $\mathbf s_g=\mathbf W_g\mathbf h$,
$\mathbf s_u=\mathbf W_u\mathbf h$, and
$\mathbf f(\mathbf h)=\mathbf W_d[
\boldsymbol\phi(\mathbf s_g)\odot\mathbf s_u]$, where
$\boldsymbol\phi$ is element-wise SiLU. The product and chain rules give
\begin{align}
 \mathrm d\mathbf f &=(\mathrm d\mathbf W_d) [\boldsymbol\phi(\mathbf s_g)\odot\mathbf s_u] +\mathbf
 W_d\bigl[ (\mathrm d\boldsymbol\phi(\mathbf s_g))\odot\mathbf s_u +\boldsymbol\phi(\mathbf
 s_g)\odot\mathrm d\mathbf s_u\bigr]
 \nonumber\\
 &=(\mathrm d\mathbf W_d) [\boldsymbol\phi(\mathbf s_g)\odot\mathbf s_u] +\mathbf W_d\bigl[
 \boldsymbol\phi'(\mathbf s_g)\odot ((\mathrm d\mathbf W_g)\mathbf h +\mathbf W_g\mathrm d\mathbf
 h)\odot\mathbf s_u  +\boldsymbol\phi(\mathbf s_g)\odot ((\mathrm d\mathbf W_u)\mathbf h +\mathbf
 W_u\mathrm d\mathbf h)\bigr].
 \nonumber
\end{align}
Composing these differentials with the remaining decoder operations
gives the logit derivatives in Proposition~\ref{prop:mols_structure}.

\subsection{Proof of Lemma~\ref{lem:mols_occurrence}}
\label{app:mols_occurrence}

\begin{molslemma}[Occurrence-conditioned embedding SNR]
\label{lem:mols_occurrence}
At a fixed parameter state, let $g$ be a raw loss-gradient coordinate
in an untied input embedding row, with $\mathbb E[g^2]<\infty$.
Let $A$ be the event that its index occurs in the input.
Suppose $0<\rho=\mathbb P(A)<1$ and
$\nu_A=\operatorname{Var}(g\mid A)>0$.
Writing $\mu_A=\mathbb E[g\mid A]$ and $s_A=\mu_A^2/\nu_A$, we have
\begin{equation}
 s=\frac{\rho s_A}{1+(1-\rho)s_A},
 \label{eq:mols_occurrence_snr}
\end{equation}
with $s\leq\rho s_A$ and $s\leq\rho/(1-\rho)$.
For an embedding tensor with vocabulary size $V$, assume finite
second moments for its gradient coordinates and occurrence
probabilities $\rho_r$ with $\rho_{\max}<1$.
If each input has at most $L$ indices, then
\begin{equation}
 S_{\mathrm{Emb}} \leq\frac1V\sum_{r=1}^V\frac{\rho_r}{1-\rho_r} \leq\frac{\mathbb E[K(\mathbf
 x)]}{V(1-\rho_{\max})} \leq\frac{L}{V(1-\rho_{\max})}.
 \nonumber
\end{equation}
\end{molslemma}

\begin{proof}
The untied lookup gives $g=0$ on $A^c$. Conditioning on occurrence,
\begin{align}
 \mathbb E[g] &=\rho\mathbb E[g\mid A]+(1-\rho)\mathbb E[g\mid A^c] =\rho\mu_A, \qquad \mathbb
 E[g^2] =\rho\mathbb E[g^2\mid A] =\rho(\nu_A+\mu_A^2),
 \nonumber\\
 \operatorname{Var}(g) &=\rho(\nu_A+\mu_A^2)-\rho^2\mu_A^2 =\rho\nu_A+\rho(1-\rho)\mu_A^2,
 \nonumber\\
 s&=\frac{\rho^2\mu_A^2} {\rho\nu_A+\rho(1-\rho)\mu_A^2} =\frac{\rho s_A}{1+(1-\rho)s_A}.
 \nonumber
\end{align}
The denominator is at least $1$ and, for $s_A>0$, at least
$(1-\rho)s_A$. These yield $s\leq\rho s_A$ and
$s\leq\rho/(1-\rho)$; both bounds also hold at $s_A=0$.
If $\nu_A=0$ but $\mu_A\neq0$, the variance expression gives
$s=\rho/(1-\rho)$. If $\rho=0$ or $\nu_A=\mu_A^2=0$,
the gradient is zero and contributes score zero.

Every coordinate in row $r$ shares its occurrence event.
Let $\mathcal A(\mathbf x)$ be the set of appearing indices.
Averaging the coordinate bounds yields
\begin{align}
 S_{\mathrm{Emb}} &=\frac1{Vd}\sum_{r=1}^V\sum_{k=1}^d s_{E,rk}
 \leq\frac1{Vd}\sum_{r=1}^V\sum_{k=1}^d \frac{\rho_r}{1-\rho_r}
 =\frac1V\sum_{r=1}^V\frac{\rho_r}{1-\rho_r}
 \nonumber\\
 &\leq\frac{\sum_r\rho_r}{V(1-\rho_{\max})} \overset{\mathrm{(a)}}{=} \frac{\mathbb
 E[\sum_r\mathbf1_{\{r\in\mathcal A(\mathbf x)\}}]} {V(1-\rho_{\max})} =\frac{\mathbb E[K(\mathbf
 x)]}{V(1-\rho_{\max})} \leq\frac{L}{V(1-\rho_{\max})}.
 \nonumber
\end{align}
Equality (a) uses linearity of expectation; the last bound uses
$K(\mathbf x)\leq L$. The proof is completed.
\end{proof}

\subsection{Proof of Lemma~\ref{lem:mols_context_mixing}}
\label{app:mols_context_mixing}

\begin{molslemma}[Context aggregation]
\label{lem:mols_context_mixing}
Let $g$ be a finite-second-moment gradient coordinate at a fixed
parameter state. Set $\mu(\mathbf x)=\mathbb E[g\mid\mathbf x]$,
$\nu(\mathbf x)=\operatorname{Var}(g\mid\mathbf x)$,
$A=\mathbb E[\mu(\mathbf x)^2]>0$, and
$B=\mathbb E[\nu(\mathbf x)]>0$.
Define $c=(\mathbb E[\mu(\mathbf x)])^2/A$ and $\Lambda=A/B$.
Then $0\leq c\leq1$ and
\begin{equation}
 \operatorname{SNR}(g) =\frac{cA}{B+(1-c)A} =\frac{c\Lambda}{1+(1-c)\Lambda}.
 \label{eq:mols_context_snr}
\end{equation}
\end{molslemma}

\begin{proof}
By total expectation and Jensen's inequality,
$(\mathbb E[g])^2=(\mathbb E[\mu(\mathbf x)])^2
\leq\mathbb E[\mu(\mathbf x)^2]=A$, so
$0\leq c\leq1$ and $(\mathbb E[g])^2=cA$.
The total-variance identity then gives
\begin{align}
 \operatorname{Var}(g) &=\mathbb E[\operatorname{Var}(g\mid\mathbf x)] +\operatorname{Var}(\mathbb
 E[g\mid\mathbf x]) =B+\mathbb E[\mu(\mathbf x)^2] -(\mathbb E[\mu(\mathbf x)])^2 =B+(1-c)A,
 \nonumber\\
 \operatorname{SNR}(g) &=\frac{cA}{B+(1-c)A} =\frac{c(A/B)}{1+(1-c)(A/B)}
 =\frac{c\Lambda}{1+(1-c)\Lambda}.
 \nonumber
\end{align}
If $A=0$, then $\mu(\mathbf x)=0$ almost surely, giving score zero
whenever the variance is positive. If $B=0$, the first fraction in
Eq.~(\ref{eq:mols_context_snr}) remains valid whenever its denominator
is positive. An identically zero gradient contributes score zero.
The proof is completed.
\end{proof}

\paragraph{Module aggregation and mini-batches}
Applying Lemma~\ref{lem:mols_context_mixing} to each coordinate gives
\begin{equation}
 S_m^t =\sum_i\omega_{m,i} \frac{c_{m,i}^t\Lambda_{m,i}^t} {1+(1-c_{m,i}^t)\Lambda_{m,i}^t},
 \nonumber
\end{equation}
with the zero cases interpreted as in the lemma.
The coefficients $c_{m,i}^t$ retain cancellation of conditional means
across contexts, rather than averaging conditional scores
$S_m^t(\mathbf x)$.

For $B_{\mathrm{mb}}$ independent copies of a complete gradient
observation at the same state, their average $\overline g$ has
$\mathbb E[\overline g]=\mathbb E[g]$ and
$\operatorname{Var}(\overline g)=\operatorname{Var}(g)/B_{\mathrm{mb}}$.
Thus $\operatorname{SNR}(\overline g)
=B_{\mathrm{mb}}\operatorname{SNR}(g)$, preserving relative population
scores when the numerical stabilizer is omitted.
For a sequence with $T$ equally weighted predicted positions,
write $g_{m,i}=T^{-1}\sum_{\ell=1}^Tg_{m,i,\ell}$. Then
\begin{align}
 \mathbb E[g_{m,i}] &=\frac1T\sum_{\ell=1}^T\mathbb E[g_{m,i,\ell}],
 \nonumber\\
 \operatorname{Var}(g_{m,i}) &=\mathbb E\left[ \left(\frac1T\sum_{\ell=1}^T (g_{m,i,\ell}-\mathbb
 E[g_{m,i,\ell}])\right)^2\right] =\frac1{T^2}\sum_{\ell,k=1}^T
 \operatorname{Cov}(g_{m,i,\ell},g_{m,i,k}).
 \nonumber
\end{align}
Calibration uses the complete mini-batch loss gradient, retaining
these cross-position covariance terms. This is the observation unit
in Corollary~\ref{cor:mols_finite_sample}.